\documentclass[preprint,review,10pt]{elsarticle}

\usepackage{graphicx}
\usepackage{booktabs}
\usepackage{subfigure}
\usepackage{caption}
\usepackage{algorithm}
\usepackage{algpseudocode}
\usepackage{array}
\usepackage{multirow}
\usepackage{enumitem}
\usepackage{amssymb}
\usepackage{mathrsfs}
\usepackage{slashbox}
\usepackage{tabularx}
\usepackage{amsmath}
\usepackage{setspace}
\usepackage{makecell}
\usepackage{bm}
\usepackage{makecell}
\usepackage{adjustbox}
\usepackage[modulo]{lineno}
\usepackage{geometry}
\usepackage{color}

\usepackage[breaklinks=true,hyperindex=true,colorlinks=true,citecolor=green, pdfauthor=author]{hyperref}

\journal{Pattern Recognition}

\newcommand{\etal}{\textit{et al}. }

\newcounter{RomanNumber}

\begin{document}
\begin{frontmatter}

\title{ Integrating Information Theory and Adversarial Learning \\ for Cross-modal Retrieval}

\author[label1]{Wei Chen}
\author[label2]{Yu Liu}
\author[label1]{Erwin M. Bakker}
\author{Michael S. Lew\corref{cor2}\fnref{label1}}
\cortext[]{Yu  Liu now is an associate professor at international school of information science and engineering, Dalian University of Technology, China  }
\cortext[cor2]{Corresponding author}
\ead{m.s.k.lew@liacs.leidenuniv.nl}
\address[label1]{LIACS, Leiden University, Leiden, 2333 CA, The Netherlands}
\address[label2]{ESAT-PSI, KU Leuven, Heverlee-Leuven, 3001, Belgium}

\begin{abstract}
Accurately matching visual and textual data in cross-modal retrieval has been widely studied in the multimedia community. To address these challenges posited by the heterogeneity gap and the semantic gap, we propose integrating Shannon information theory and adversarial learning. In terms of the heterogeneity gap, we integrate modality classification and information entropy maximization adversarially. For this purpose, a modality classifier (as a discriminator) is built to distinguish the text and image modalities according to their different statistical properties. This discriminator uses its output probabilities to compute Shannon information entropy, which measures the uncertainty of the modality classification it performs. Moreover, feature encoders (as a generator) project uni-modal features into a commonly shared space and attempt to fool the discriminator by maximizing its output information entropy. Thus, maximizing information entropy gradually reduces the distribution discrepancy of cross-modal features, thereby achieving a domain confusion state where the discriminator cannot classify two modalities confidently.  To reduce the semantic gap, Kullback-Leibler (KL) divergence and bi-directional triplet loss are used to associate the intra- and inter-modality similarity between features in the shared space. Furthermore, a regularization term based on KL-divergence with temperature scaling is used to calibrate the biased label classifier caused by the data imbalance issue. Extensive experiments with four deep models on four benchmarks are conducted to demonstrate the effectiveness of the proposed approach.

\end{abstract}


\end{frontmatter}

\vspace{1cm}

\section{Introduction}
\label{sec:introduction}

Semantic information that helps us understand the world usually comes from different modalities such as video, audio, and text. Namely, the same concept can be presented in different ways. Therefore, it is possible to search semantically-relevant samples (\emph{e.g.} images) from one modality when given a query item from another modality (\emph{e.g.} text). With the increasing amount of multimodal data available, more efficient and accurate retrieval methods are still in demand in the multimedia community.

Deep learning methods can effectively embed features from different modalities into a commonly shared space, and then measure the similarity between these embedded features. To date, the ``heterogeneity gap'' \cite{wang2016comprehensive} and the ``semantic gap'' \cite{li2016socializing} are still challenges to be addressed for cross-modal retrieval. Since the data in different modalities are described by different statistical properties, the heterogeneity gap characterizes the difference between feature vectors from different modalities that have similar semantics but are distributed in different spaces. Similarities between these feature vectors are not well associated so that these vectors are not directly comparable, leading to inconsistent distributions. The semantic gap characterizes the difference between the high-level user perception of the data and the lower-level representations of the data by the computer (\emph{i.e.} pixels or symbols). To achieve better retrieval performance, it is essential to address these gaps for associating the similarity between cross-modal features in the shared space.

To capture the semantic correlations between cross-modal features, many approaches have been proposed in recent years. Some approaches focus on designing effective structures from a deep networks perspective. For instance, graph convolutional networks are employed to model the dependencies within visual or textual data \cite{angelou2019graph}. Other approaches focus on designing similarity constraint functions from a deep features perspective. For example, bilinear pooling-based methods are applied to align image and text features to then accurately capture inter-modality semantic correlations. In other examples, coordinated representation learning methods \cite{baltruvsaitis2018multimodal}, such as ranking loss \cite{faghri2017vse, wang2017adversarial} and cycle-consistency loss \cite{liu2019cyclematch} are widely used to preserve similarity between cross-modal features. These constraint functions mainly aim at reducing the semantic gap by focusing on the similarity between two-tuple or three-tuple samples. However, they might not directly mitigate the heterogeneity gap caused by the inconsistent feature distributions in the different spaces. 

\begin {figure}[!t]
\centering
 {
   \includegraphics[width=0.6\columnwidth]{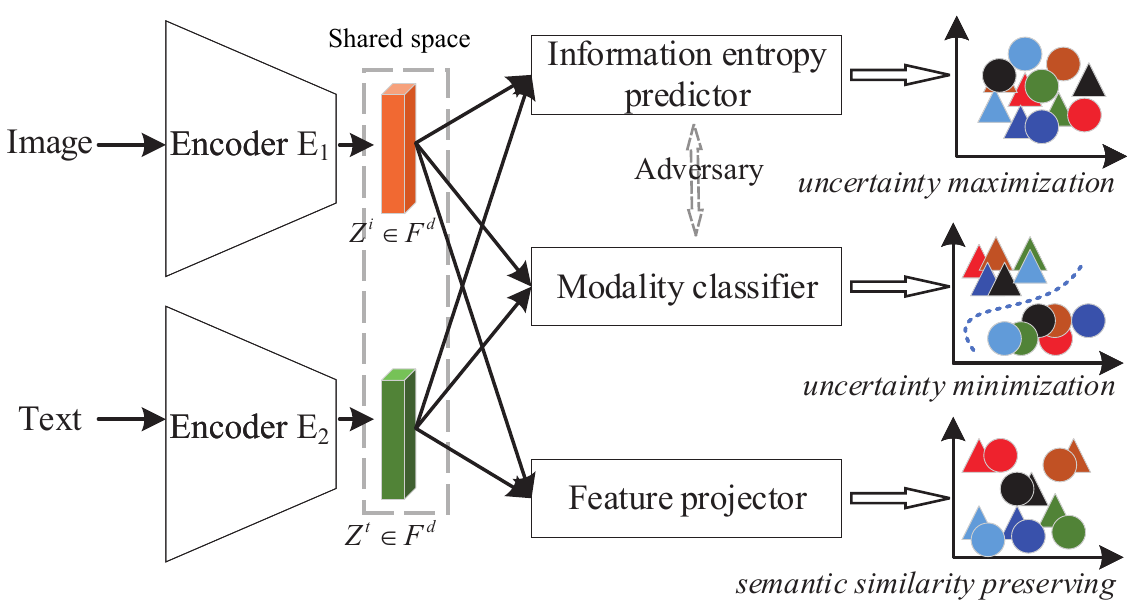} 
 }
\caption{Conceptual diagram of combining information theory and adversarial learning for cross-modal retrieval. The features $ Z^{i}\in F^{d}$  and $Z^{t}\in F^{d}$ with dimension $d$ for image-text pairs are extracted using deep neural networks. Shape indicates modality and color denotes pair-wise similarity information. The modality classifier aims to classify the text and image modalities, thereby minimizing the uncertainty of modality classification it performs (measured by Shannon information entropy). Conversely, the feature encoders project uni-modal features into a commonly shared space and attempt to fool this classifier by maximizing its uncertainty of modality classification, which is computed by the information entropy predictor. The modality classifier and the information entropy predictor are combined in an adversarial manner to reduce the heterogeneity gap. If the classifier's uncertainty is maximized, features $ Z^{i}$ and $ Z^{t}$ are intertwined into a domain confusion state where this classifier cannot confidently determine which modality each input feature ($ Z^{i}$ or $ Z^{t}$) belongs to. Namely, this classifier becomes least-confident on its classification results. This process of adversarial combining is introduced in Section \ref{Information_Theory_Meets_Adversarial_Learning} and Section \ref{Combining_Information_Theory_into_Discriminator}. Furthermore, the feature projector aims to associate the semantic similarity by using pair-wise objective functions such as bi-directional triplet loss.}
\label{ProblemSetting}
\end {figure}

\subsection{Motivations}
\label{Our_motivation}

Considering the limitations of similarity constraint functions, we propose a new method to perform cross-modal retrieval from two aspects. First, we reduce the heterogeneity gap by integrating Shannon information theory \cite{shannon1948mathematical} with adversarial learning, in order to construct a better embedding space for cross-modal representation learning. Second, we combine two loss functions, including Kullback-Leibler divergence loss and bi-directional triplet loss, to preserve semantic similarity during the feature embedding procedure, thereby reducing the semantic gap.

To do this, we combine the information entropy predictor and the modality classifier in an adversarial manner. Information entropy maximization and modality classification are two processes trained with competitive goals. Since the image is a 3-channel RGB array while the text is often symbolic, uni-modal features extracted from image or text data are characterized by different statistical properties, which can be used to distinguish the original modalities these features belong to. As a result, when these features in the shared space are correctly classified into their original modalities with high confidence, then their feature distributions convey less information content, and the modality classifier performs modality classification with lower uncertainty. In contrast, when cross-modal features become modality-invariant and show their commonalities, these features cannot be classified into the modality they originally belong to. In this case, the feature distributions in the shared space conveys more information content and higher modality uncertainty.

According to Shannon's information theory \cite{shannon1948mathematical}, we can measure the modality uncertainty in the shared space by computing information entropy. This basic proportional relation provides the principle to mitigate the heterogeneity gap. For this purpose, we integrate modality uncertainty measurement into cross-modal representation learning. As shown in Figure \ref{ProblemSetting}, a modality classifier (in the following we call it a \textit{discriminator}) is devised to classify image and text modality, rather than perform a ``true/false'' binary classification. This discriminator also provides its output probabilities to calculate the information entropy of the cross-modal feature distributions. At the start of training, the discriminator can classify images and text modalities with high confidence due to their different statistical properties. In contrast, the feature encoders (in the following we call it a \textit{generator}) project features into a shared space and attempt to fool the discriminator and make it perform an incorrect modality classification until features in the shared space are fused heavily into a confusion state, maximizing the modality uncertainty.

On the basis of this heavily-fused state, we further use similarity constraints on the feature projector to reduce the semantic gap. Specifically, Kullback-Leibler (KL) divergence loss is used to preserve semantic correlations between image and text features by using instance labels as supervisory information. More importantly, we consider the issue of data imbalance and introduce a regularization term based on KL-divergence with temperature scaling to calibrate the biased label classifier. Afterwards, we adopt the commonly used bi-directional triplet loss and instance label classification loss (\emph{i.e.} categorical cross-entropy loss) to achieve good retrieval performance.
 
\subsection{Our Contributions}

Our contributions can be summarized three-fold as follows:

First, we combine information theory and adversarial learning into an end-to-end framework. Our work is the first to explore information theory in reducing the heterogeneity gap for cross-modal retrieval. This method is beneficial for constructing a shared space for further learning commonalities between cross-modal features, which can be used for tasks in other modalities, such as video-text matching.

Second, we introduce a regularization term based on KL-divergence with temperature scaling to address the issue of data imbalance, which calibrates biased label classifier training and guarantees the accuracy of instance label classification. To the best of our knowledge, we are not aware of any prior use in the context of addressing imbalance issues on retrieval datasets.

Third, we use bi-directional triplet loss to constrain intra-modality semantics. Aside from these intra-modality constraints, we also consider optimizing inter-modality similarity. We use the instance labels to construct a supervisory matrix. This matrix regularizes the semantic similarity between the projected image (or text) features and text (or image) features by minimizing KL-divergence. This inter-modality constraint is more effective since it focuses on all the projected cross-modal feature distributions in a mini-batch.

The rest of paper is organized as follows. Related work is reviewed in Section \ref{related_work}. We give definitions and a theoretical analysis for the proposed method in Section \ref{Information_Theory_Meets_Adversarial_Learning}. We present the specific components for implementation including network structures, objective functions, and optimization in Section \ref{Implementation_and_Optimization}. We test the proposed method on four datasets, and the results are reported in Section \ref{Experiments}. Finally, the conclusions are given in Section \ref{Conclusion}.

\section{Related Work}
\label{related_work}

\subsection{Cross-modal Representation Learning and Matching}

Preserving the similarity between cross-modal features should consider two aspects: inter-modality and intra-modality. Supervision information (\emph{e.g.} class label or instance label), if available, is beneficial for learning features from these two aspects. Preserving feature similarity can be realized by using methods such as joint representation learning and coordinated representation learning \cite{baltruvsaitis2018multimodal}. Joint representation learning methods project the uni-modal features into the shared space using straightforward strategies such as feature concatenation, summation, and inner product. Subsequently, more complicated bilinear pooling methods, such as multimodal compact bilinear (MCB) pooling, are proposed to explore the semantic correlations of cross-modal features. To regularize the joint representations, deep networks are commonly trained by using objective functions, such as regression-based loss \cite{mhiri2019word, wang2020joint}. 

Coordinated representation learning methods process image and text features separately but impose them under certain similarity constraints \cite{baltruvsaitis2018multimodal}. In general, these constraints can be categorized into classification-based and verification-based methods in supervised scenarios. In terms of classification-based methods, both image and text features are used to make a label classification by using categorical cross-entropy loss function. Because a paired image-text input has the same class label, their features can be associated in the shared space. However, classification-based methods cannot preserve the similarity between inter-modality features well because the similarity between image and text features is not directly regularized.

Verification-based methods, based on metric learning, are proposed to further optimize inter-modality feature learning. Given a similar (or dissimilar) image-text pair, their corresponding features should be verified as similar (or dissimilar). Therefore, the goal of deep networks is to push features of similar pairs closer, while keeping features of dissimilar pairs further apart. Verification-based methods include pair-wise constraints and triplet constraints, which focus on inferring the matching scores of image-text feature pairs \cite{wang2020joint}.

Triplet constraints optimize the distance between positive pairs to be smaller than the distance between negative pairs by a margin. They can capture both intra-modality and inter-modality semantic correlations. For example, bi-directional triplet loss has been employed to optimize image-to-text and text-to-image ranking \cite{wang2017adversarial}. Although triplet constraints are widely used for cross-modal retrieval, the difficulties are in the mining strategy for negative pairs and the selection of a margin value, which are usually task-specific and empirically selective.

\subsection{Adversarial Learning for Cross-modal Retrieval}

The afore-mentioned joint and coordinated representation learning approaches focus on two-tuple or three-tuple samples, which may be insufficient for achieving overall good retrieval performance. Adversarial learning, as an alternative method, has shown its powerful capability for modeling feature distributions and learning discriminative representations between modalities when deep networks are trained with competitive objective functions \cite{wang2017adversarial,wu2020modality}.

Recent progress in using adversarial learning for cross-modal retrieval can be categorized as feature-level and loss function-level discriminative models. 

From a feature-level perspective, it is possible to preserve semantic consistency by performing a min-max game between inter-modality feature pairs  \cite{wang2017adversarial}. A straightforward way is to build a discriminator, making a ``true/false'' classification between image features (regarded as true), corresponding matched text features (regarded as fake), and unmatched image features from other categories (also regarded as fake) \cite{wang2017adversarial}. Alternatively, a cross-modal auto-encoder can be combined to generate features for another modality. For example, a generator attempts to generate image features from textual data and then regards them as true, while for a discriminator, image features extracted from original images and these from the generated ``images'' are labeled as true and fake, respectively. The adversarial training explores the semantic correlations of cross-modal representations. Intra-modality discrimination also can be considered in cross-modal adversarial learning, forcing the generator to learn more discriminative features. In this case, the discriminator tends to discriminate the generated features from its original input.

From a loss function-level perspective, instead of making a binary classification (\emph{i.e.} true or fake), adversarial learning is designed to train two groups of loss functions or two processes with competitive goals. This idea is applied in recent work for cross-modal retrieval \cite{wang2017adversarial, wu2020modality}. To be specific, a feature projector is trained to generate modality-invariant representations in the shared space, while a modality classifier is constructed to classify the generated representations into two modalities. Similarly, in this paper, we combine two networks and train them with two competitive goals.

\subsection{Information-theoretical Feature Learning}

As mentioned before, feature vectors from different modalities are distributed in different spaces, resulting in the heterogeneity gap, which affects the accuracy of cross-modal retrieval. Therefore, it becomes essential to reduce feature distribution discrepancies and thereby reduce the heterogeneity gap. The solution for this is to measure and then minimize distribution discrepancy. For example, distribution disparity of cross-modal features can be characterized by Maximum Mean Discrepancy (MMD), which is a differentiable distance metric between distributions. However, MMD suffers from sensitive kernel bandwidth and weak gradients during training.

Information-theoretical based methods are used to measure the differences of feature distributions and learn better cross-modal features. As an example, the cross-entropy loss function is widely used to estimate the errors between inference probabilities and ground-truth labels where the gradients are computed according to the errors. Once the gradients are computed, deep networks can further update their parameters via the back-propagation algorithm. KL-divergence (also called relative entropy) is another popular criterion to characterize the difference between two probability distributions. Minimizing the difference is beneficial for retaining the semantic similarity between features. For example, Zhang \etal \cite{zhang2018deep} employ the KL-divergence to measure the similarity between projected features and supervisory information. 

Recently, Shannon information entropy \cite{shannon1948mathematical} has been used for performing tasks such as semantic segmentation \cite{vu2019advent} and cross-modal hash retrieval \cite{chen2019domain}. These studies indicate that Shannon entropy can be used for multimodal representation learning by estimating uncertainty \cite{shannon1948mathematical}. Take generative adversarial networks as an example: if the generator makes image features and text features close and minimizes their discrepancy, then the discriminator will become less-certain or under-confident, \emph{i.e.}, having a high information entropy to predict which modality each feature comes from. We applied this principle in our previous work \cite{chen2019domain} to design an objective function to maximize the domain uncertainty over cross-modal hash codes in a commonly shared space. Deep networks trained by using information entropy construct a domain confusion state where the heterogeneity gap can be effectively reduced. On the basis of this state, other loss functions, such as ranking loss, can be further applied to regularize feature similarity.

\section{Proposed Approach}

\subsection{Problem Formulation}
\label{Problem_Formation}

We consider a supervised scenario for cross-modal retrieval. Denote $X^{i}$ as the input images and the corresponding descriptive sentences as $ X^{t}$. Each image and its descriptive sentences have the same instance label $ Y $. Therefore, we can organize an input pair ($x^{i}$, $x^{t}$, $y$) to train a deep network. To be specific, feature encoders $\mathit{E}_{1}(\cdot;\boldsymbol \theta_{E_{1}})$ and  $\mathit{E}_{2}(\cdot;\boldsymbol \theta_{E_{2}})$ extract image and text features, respectively, and then further embed these uni-modal features into a shared space by using non-shared sub-networks. The embedded features with dimension $d$ are denoted as $ Z^{i}=\mathit{E}_{1}(X^{i};\boldsymbol \theta_{E_{1}})$ and $ Z^{t}=\mathit{E}_{2}(X^{t};\boldsymbol \theta_{E_{2}}) $, $ Z^{i}$,$ Z^{t} \in R^{d}$. Note that the parameters in the non-shared sub-networks for uni-modal image and text feature embedding have been included into $\boldsymbol \theta_{E_{1}}$ and $\boldsymbol \theta_{E_{2}}$, respectively. The goal is to train a deep network to make the embedded features $ Z^{i}$ and $ Z^{t} $ modality-invariant and semantically discriminative, improving the retrieval accuracy.

As shown in Figure \ref{ProblemSetting}, the networks $\mathit{E}_{1}$, $\mathit{E}_{2}$, and the information entropy predictor act as a generator, while the modality classifier acts as a discriminator. The training of the generator and the discriminator is formulated as an interplay min-max game to mitigate the heterogeneity gap. The feature projector attempts to preserve feature similarity under several constraints, which are introduced in Section \ref{KL_divergence_for_Similarity_Preserving}, \ref{Instance_Label_Classification}, and \ref{Bi_directional_Triplet_Constraint}. 

\subsection{Integrating Information Theory and Adversarial Learning}
\label{Information_Theory_Meets_Adversarial_Learning}

\subsubsection{Information Entropy and Modality Uncertainty}
\label{Information_Entropy_and_Modality_Uncertainty}

\begin {figure}[!t]
\centering
  {     
   \includegraphics[width=0.6\columnwidth]{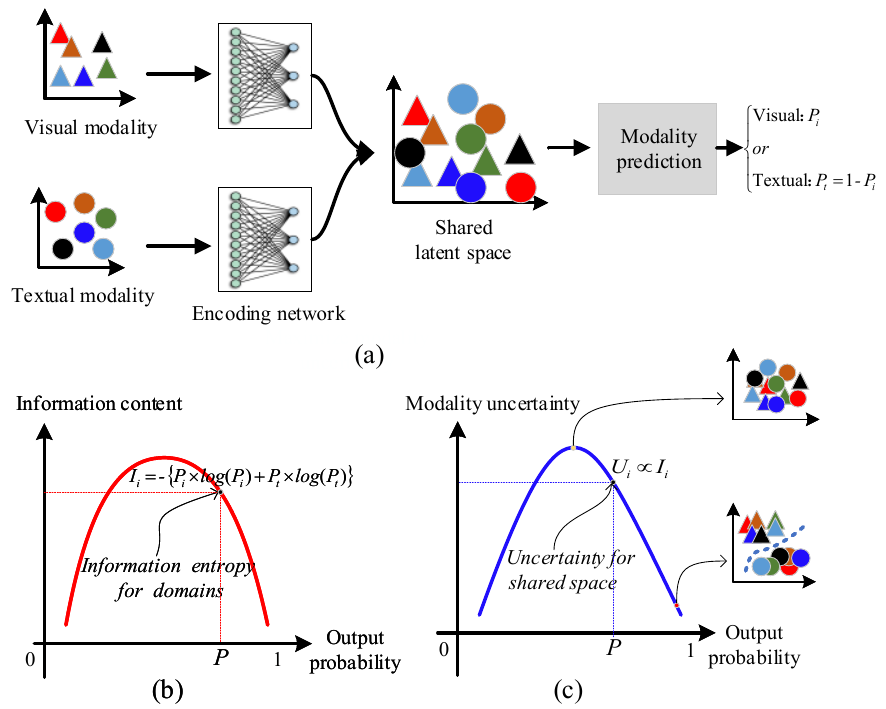}  
 }  
 \caption{ (a): Image and text features are further embedded into a shared space via non-shared encoding sub-networks. The modality uncertainty can be predicted by using the output classification probabilities from a predictor. (b): Relationship between output probabilities and information content. The more uncertain the shared space, the more information content it conveys. (c): Relationship between modality uncertainty and output probabilities for each modality. When probabilities predicted for two modalities are identical, the shared space is intertwined into a domain confusion state (\emph{i.e.} most uncertain). If one modality is identified with a higher probability (closer to 1) while another with a lower probability (closer to 0), the domain confusion state is not achieved. } 
\label{Figure2_illutration_V3}
\end {figure}

Image features can be extracted from convolutional neural networks, while text features can be extracted from sequential networks. These feature vectors from different modalities have similar semantics but are distributed in different spaces. Their similarities in the different spaces are not well associated so that these feature vectors are not directly comparable. Hence, it is required to further embed them into a shared space (\emph{i.e.} $ Z^{i} $ and $ Z^{t} $ in Figure \ref{ProblemSetting}). Uni-modal features are characterized by different statistical properties. Therefore, as shown in Figure \ref{Figure2_illutration_V3}(a), it is possible to identify a feature in the shared space coming from a visual modality with higher probability $ P_{i} $ (more certain classification) than coming from a textual modality with lower probability $ P_{t}\!=\!\!1\!-\!P_{i} $ (less certain classification). In other words, these cross-modal features are not intertwined heavily. As a result, the domain confusion state is not achieved. Conversely, if a given feature can not be distinguished which modality this feature originally comes from, it indicates that this feature has identical probability ($ P_{i}\!=\!P_{t}$) coming from each modality. In this case, the shared space has highest uncertainty and the cross-modal features are intertwined into a domain confusion state, which corresponds to highest information content. We use information entropy \cite{shannon1948mathematical} to measure the uncertainty of the shared space. Figure \ref{Figure2_illutration_V3}(b) illustrates that two modalities with an equal probability leads to the highest Shannon information entropy and thus information content.

Modality uncertainty refers to the unreliability of classification that the discriminator classifies image features and text features into two modalities. It is proportional to Shannon information entropy \cite{shannon1948mathematical}, as shown in Figure \ref{Figure2_illutration_V3}(c). Based on this observation \cite{chen2019domain}, we design the discriminator to measure its output modality uncertainty by using information entropy as a criterion. Maximizing information entropy means that the discriminator becomes least-confident in classifying the original modality of image and text features, resulting in the greatest reduction of the heterogeneity gap.

\subsubsection{Adversarial Learning and Information Entropy}

To make cross-modal features modality-invariant, we devise a generator and a discriminator, as shown in Figure \ref{ProblemSetting}. The discriminator performs modality classification to identify visual modality and textual modality based on cross-modal features. Following \cite{wang2017adversarial}, we define the modality label as $ Y^{*}_{c}$ for these two modalities (for visual modality  $ * = i $ and textual modality $ *=t$). Using output probabilities of the discriminator, we can compute cross-entropy loss to realize modality classification \cite{wang2017adversarial}. Once the network convergences under the constraint of this loss function, visual modality and textual modality are clearly identified and classified, thereby minimizing the modality uncertainty.

Conversely, the generator is designed to maximize the modality uncertainty over the cross-modal feature distributions. To achieve this, the generator learns modality-invariant features to fool the discriminator, maximizing the uncertainty of modality classification the discriminator performs. If the modality uncertainty is maximized, the discriminator is most likely to make an incorrect modality classification and be least-confident about its classification results. In this case, cross-modal features are intertwined into a domain confusion state and become indistinguishable.

To this end, we explore the ways to integrate information entropy and adversarial learning into an end-to-end network, which is introduced in Section \ref{Combining_Information_Theory_into_Discriminator}. For better understanding, we also explore another combining paradigm in the Experimental Section.

\subsection{KL-divergence for Cross-modal Feature Projection }
\label{Cross_modal_feature_projection}

\begin {figure}[!t]
\centering
  {     
   \includegraphics[width=0.6\columnwidth]{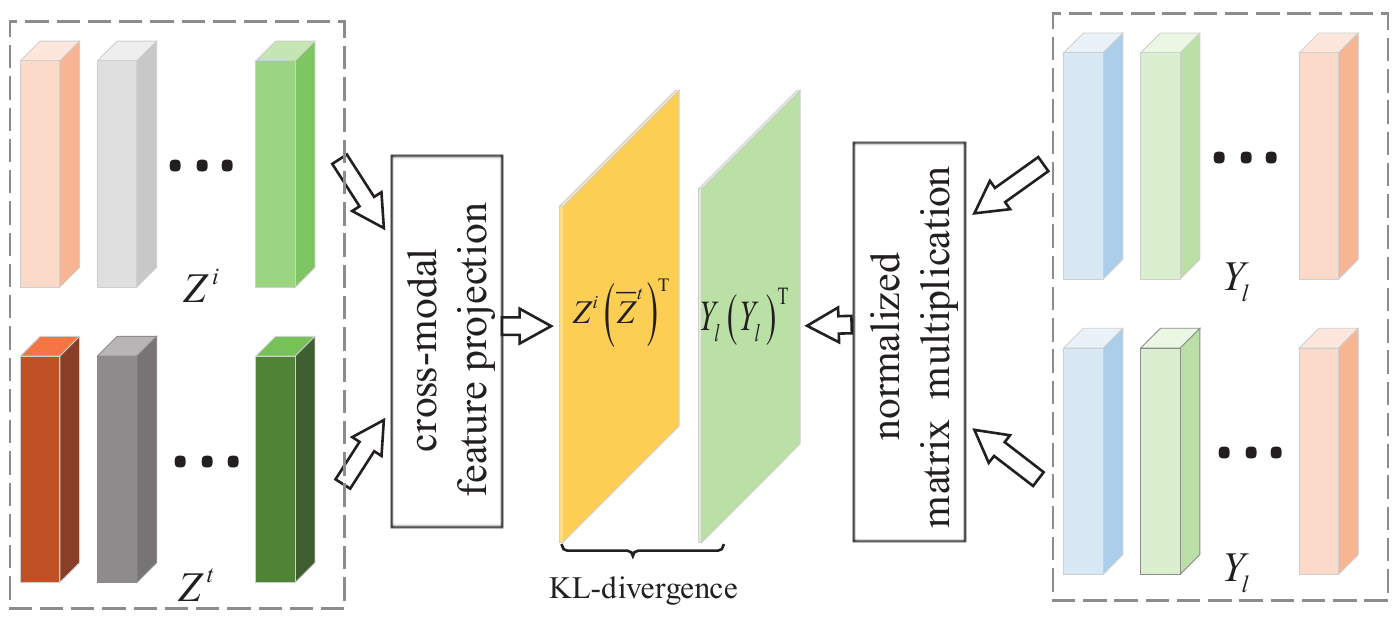}  
 }  
 \caption{ KL-divergence for cross-modal feature projection, which considers all features $ Z^{i} $ and $ Z^{t} $ in the shared space. Each paired image feature and text feature share the same instance label, indicated by the same color. The cross-modal feature projection module is critical to explore the similarity between image features and normalized text features. The projection process is formulated in Eqs. \ref{KL_projection1} and \ref{KL_projection2}.} 
\label{Figure3_cross_modal_projection}
\end {figure}

To reduce the semantic gap,  we use KL-divergence to characterize the differences between projected cross-modal features ($ Z^{i} $ and $ Z^{t} $ in Figure \ref{ProblemSetting}) and a supervisory matrix computed from their instance labels, \emph{i.e.} $ KL((f(Z^{i}, Z^{t})||f(Y^{\top}_{l},Y_{l})) $, (see Eq. \ref{KL_divergence_symmetrical_loss}). In this way, the semantic correlations among cross-modal features can be preserved. We illustrate this process in Figure \ref{Figure3_cross_modal_projection}. It is important to note that when using KL-divergence to preserve semantic correlations of cross-modal features, all positive and negative pairs in a mini-batch are considered. As for the supervisory matrix $f(Y^{\top}_{l},Y_{l})$, it is computed by using matrix multiplication and is normalized to the range from 0 and 1.

We argue that different operations to realize $f(Z^{i}, Z^{t})$ affect similarity preserving. Directly, the operation $f(\cdot)$ can be an inner product on cross-modal features $Z^{i}$ and $ Z^{t}$. However, using the inner product has some implicit drawbacks. First, when multiplying one image feature vector with all text feature vectors, the results of the inner product are not optimally comparable due to the non-normalized text features, and vice versa. Second, the angles between each image feature vector and each text feature vector, as well as their whole feature distributions, are changing when training the deep network, which makes it problematic for an inner product to measure feature similarity.

To tackle the above limitations, we adopt a cross-modal feature projection to characterize the similarity between features. The idea is related to the work in \cite{zhang2018deep}. Cross-modal feature projection is based on the same distribution and operates on the normalized features. For instance, an image feature vector, $ z^{i}_{j} \in Z^{i} $, can be projected to the distribution of a text feature vector $ z^{t}_{k} \in Z^{t} $, then each projected feature vector from image to text (termed ``$ i \to t $'') can be formulated as:

\begin{equation}
\label{Feature_projection}
\setlength\abovedisplayskip{3pt}
\setlength\belowdisplayskip{1pt}
\begin{aligned}
 \hat{z}^{i \to t}_{j} & = |z^{i}_{j}| \ast \frac{<\!\!z^{i}_{j}, z^{t}_{k}\!\!>}{|z^{i}_{j}| |z^{t}_{k}|} \ast \frac{z^{t}_{k}}{|z^{t}_{k}|}
\\
& = <\!\!z^{i}_{j} , \bar{z}^{t}_{k}\!\!> \ast \bar{z}^{t}_{k}
\end{aligned}
\end{equation}
where ``$ i $'' and ``$ t $'' represent the visual and the textual modality, respectively, ``$ j $'' and ``$ k $'' represent the index of each image feature and text feature in the shared space, respectively, $\bar{z}^{t}_{k}$ denotes the normalized feature. Therefore, the length of $ \hat{z}^{i \to t}_{j} $ is equal to $ |\hat{z}^{i \to t}_{j} | \! = \! | \!\! <\!\!z^{i}_{j}, \bar{z}^{t}_{k}\!\!> \!\!| $, and denotes the similarity between image feature $ z^{i}_{j} $ and text feature $ z^{t}_{k} $. When associating each image feature $ z^{i}_{j} $ with all text features $ Z^{t}$, we obtain all different lengths, Therefore, when projecting all image features into all text features $ Z^{t}$, we get a similarity matrix $ A_{i \to t}$, which is formulated as

\begin{equation}
\label{KL_projection1}
\setlength\abovedisplayskip{3pt}
\setlength\belowdisplayskip{1pt}
\begin{aligned}
 A_{i \to t}(Z^{i}, Z^{t}) = \displaystyle\sum_{j=1}^{N}\!\sum_{k=1}^{N}|\!\!<\!\!z^{i}_{j} , \bar{z}^{t}_{k}\!\!>\!\!| = Z^{i}(\bar{Z}^{t})^{\top}
\end{aligned}
\end{equation}

 Similarly, if projecting all text features into all image features $ Z^{i}$, we obtain another similarity matrix $ A_{t \to i}$:

\begin{equation}
\label{KL_projection2}
\setlength\abovedisplayskip{3pt}
\setlength\belowdisplayskip{1pt}
\begin{aligned}
 A_{t \to i}(Z^{t}, Z^{i}) = \displaystyle\sum_{k=1}^{N}\!\sum_{j=1}^{N}|\!\!<\!\!z^{t}_{k} , \bar{z}^{i}_{j}\!\!>\!\!| = Z^{t}(\bar{Z}^{i})^{\top}
\end{aligned}
\end{equation}

In the above two equations, $ Z^{i} $ and $ Z^{t} $ represent the cross-modal features from two modalities. $ N $ is the number of samples in a mini-batch. These two similarity matrices are normalized by a softmax function. Afterwards, we use KL-divergence to characterize the difference between the normalized matrices and the supervisory matrix, \emph{i.e.} $ KL((f(Z^{i}, Z^{t})||f(Y^{\top}_{l},Y_{l})) $. The specific objective function is introduced in Section \ref{KL_divergence_for_Similarity_Preserving}.

\section{Implementation and Optimization}
\label{Implementation_and_Optimization}

 We introduce the implementation and optimization of our proposed approach in this section. We employ four convolutional neural networks such as ResNet-152 \cite{he2016deep} and MobileNet \cite{howard2017mobilenets} to obtain image features and a Bi-directional LSTM (Bi-LSTM) \cite{graves2005bidirectional} to extract text features. All the extracted image and text features are uni-modal. Later, we borrow the protocols of non-shared encoding sub-networks (fully-connected layers) in \cite{zhang2018deep} to get the cross-modal features $ Z^{i} $ and $  Z^{t}$.

Once the cross-modal features are obtained, we use the proposed algorithm to train the networks based on the above theoretical analysis. The algorithm includes combining information entropy and adversarial learning to mitigate the heterogeneity gap, and loss function terms (\emph{i.e.} KL-divergence loss, categorical cross-entropy loss, and bi-directional triplet loss) to preserve semantic correlations between cross-modal features.

\begin {figure}[!t]
\centering
 { 
   \includegraphics[width=0.6\columnwidth]{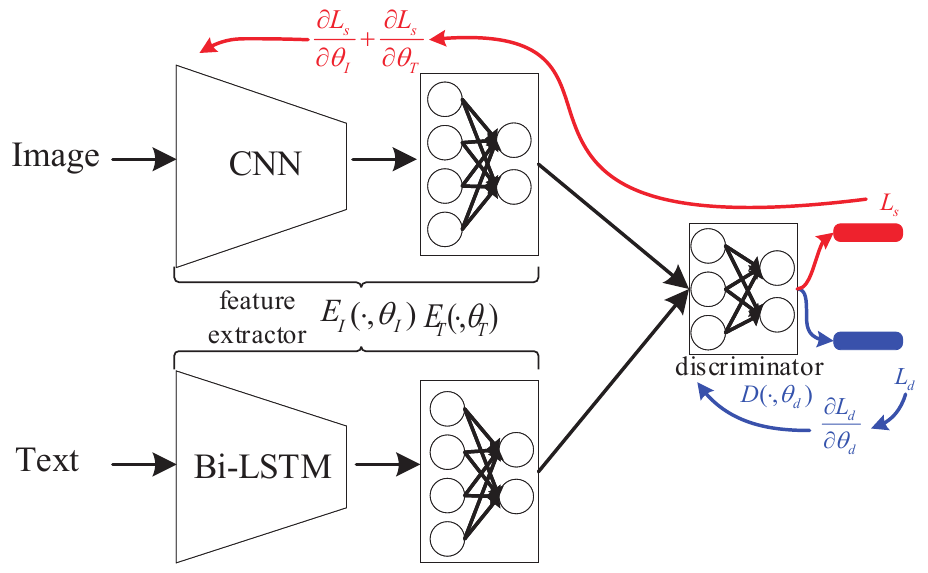} 
 }
\caption{The implementation of integrating information entropy predictor and modality classifier in Figure \ref{ProblemSetting} into a unified discriminator. Together with the feature extractors, the whole framework is in the form of generative adversarial network.  For clarity, we ignore the feature projector mentioned in Figure \ref{ProblemSetting}, which includes label classification loss, bi-directional triplet loss, and KL-divergence loss.}
\label{UnifiedDiscriminator}
\end {figure}

\subsection{Combining Information Theory with Adversarial Learning}
\label{Combining_Information_Theory_into_Discriminator}

We combine information entropy predictor and modality classifier in Figure \ref{ProblemSetting} into a unified sub-network, as shown in Figure \ref{UnifiedDiscriminator}. In this paradigm, the discriminator $D $ with parameters $ \boldsymbol \theta_{D} $ performs a modality classification and computes the Shannon information entropy. The backbone nets $ E_{1}$ and $E_{2}$ for feature extraction act as the generator $ G $. The whole structure forms a generative adversarial network. The information entropy computed from the discriminator back-propagates to the feature encoders. Specifically, when the discriminator is fixed, and its parameters are $ \boldsymbol \theta_{D}^{\star} $, then the information entropy $ H(P_{_{D}}^{\star}) = \mathbb{E}_{_{i,t}}(-P_{_{D}}^{\star} \ast log(P_{_{D}}^{\star})) $ is computed from its output probabilities $ P_{_{D}}^{\star}(D|Z^{i,t};\boldsymbol \theta_{D}^{\star})$ across the features for all classes. Based on the information entropy, we can design a negative entropy loss $L_{s} = - H(P_{_{D}}^{\star})$ (see Eq. \ref{InformationEntropy_unified}) to train the network. The gradients computed from $L_s$ update the parameters of feature extractors. The negative information entropy $L_s$ is label-free during training, and it regularizes the whole feature distribution to be modality-invariant.

The discriminator consists of some fully-connected layers. The last layer with two neurons yields probabilities that correspond to two modalities. This discriminator classifies whether the input features $ Z^{i}$ and $ Z^{t} $ are from the visual or the textual modality given the pre-defined modality label $ Y^{*}_{c}$. In contrast, the generator (\emph{i.e.} $ E_{1} $ and $ E_{2}$ ) aims at learning modality-invariant features to fool the discriminator to make an incorrect modality classification so that the generator gradually maximizes the output information entropy from the discriminator. Therefore, the learning process of the discriminator affects that of the generator in an indirect way. The objective function is calculated using the output probabilities $ P_{_{D}}(D|Z^{i,t};\boldsymbol \theta_{D})$ of the discriminator.

 For the generator  $ E_{1}$ and $E_{2}$:

\vspace{-1.0em}
\begin{equation}
\label{InformationEntropy_unified}
\begin{aligned}
 L_{s} \!&  = 
\displaystyle \frac{1}{N} \sum_{j=1}^{N}\!\sum_{m=1}^{M}\!\Big(P^{i}_{D,m}(D^{i}|Z^{i}_{j}; \boldsymbol \theta_{D} )\!\ast \!log\big(P^{i}_{D,m}(D^{i}|Z^{i}_{j}; \boldsymbol \theta_{D} )\big)
 + P^{t}_{D,m}(D^{t}|Z^{t}_{j}; \boldsymbol \theta_{D} )\!\ast \!log\big(P^{t}_{U,m}(D^{t}|Z^{t}_{j}; \boldsymbol \theta_{D} )\big)\Big) \\
& s.t. \sum_{m=1}^{M}\! P^{*}_{D,m}(D^{*}|Z^{*}_{j}; \boldsymbol \theta_{D} ) = 1, \;   P^{*}_{D,m}(D^{*}|Z^{*}_{j}; \boldsymbol \theta_{D} ) \ge 0 
\end{aligned}
\end{equation}
It is expected for the generator $G$ to maximize the information entropy $H(P_{_{D}}^{\star})$, and subsequently the modality uncertainty (see Figure \ref{Figure2_illutration_V3}). Since $L_s$ is a negative entropy ($L_{s} \! = \!\! - H(P_{_{D}}^{\star})$) to maximize $H(P_{_{D}}^{\star})$, it is minimized to optimize the parameters $\boldsymbol \theta_{E_{1}}$ and $\boldsymbol \theta_{E_{2}}$ of the generator during training.
For the discriminator $ D $, depending on the modality label $ Y^{i}_{c} $ and $ Y^{t}_{c} $ and its output probabilities $ P_{_{D}}(D|Z^{i,t};\boldsymbol \theta_{D})$, the modality classification cross-entropy loss function is formulated as:

\begin{equation}
\label{Modalityclassification_unified}
\begin{aligned}
 L_{c} \!&  = 
\displaystyle - \frac{1}{N} \sum_{j=1}^{N}\Big( Y^{i}_{c}\!\ast \!log\big(P^{i}_{D}(D^{i}|Z^{i}_{j}; \boldsymbol \theta_{D} )\big)
 +  Y^{t}_{c}\!\ast \!log\big(P^{t}_{D}(D^{t}|Z^{t}_{j}; \boldsymbol \theta_{D} )\big)\Big)
\end{aligned}
\end{equation}

$L_c$ refers to the negative cross-entropy loss of the discriminator and is minimized to clearly classify image and text features into two modalities during training. Note that the gradients calculated from term $L_s$ are only used to optimize the parameters $\boldsymbol \theta_{E_{1}}$ and $\boldsymbol \theta_{E_{2}}$ of the generator, whereas the gradients from term $L_c$ are only for optimizing the parameters $ \boldsymbol \theta_{D} $ of the discriminator, as shown in Figure \ref{UnifiedDiscriminator}. Minimizing loss $L_c$ and $L_s$ when trained iteratively will reduce the heterogeneity gap. The optimization method is straightforward, even though the gradients calculated from $L_c$ will not directly affect the parameters of the feature encoders $ E_{1} $ and $ E_{2} $. The output probabilities of the discriminator change when updating its parameters, which will affect the Shannon information entropy and affect the output features from $ E_{1} $ and $ E_{2} $ in the end.

\subsection{KL-divergence for Similarity Preserving}
\label{KL_divergence_for_Similarity_Preserving}

We also compute KL-divergence directly across $ Z^{i}$ and $ Z^{t} $ to further preserve semantic similarity. KL-divergence focuses on the projections of image and text features and is computed by  $ L_{kl} = KL((f(Z^{i}, Z^{t})||f(Y^{\top}_{l}, Y_{l})) $. Here, superscript ``$ ^{\top} $'' means matrix transpose. 
$ L_{kl}$ focuses on constraining the whole feature distributions and is complementary to the following bi-directional triplet loss function. We have introduced the process of cross-modal feature projection in Section \ref{Cross_modal_feature_projection}. Given the similarity matrices (\emph{i.e.} $ A_{i \to t}(Z^{i}, Z^{t}) $ and $  A_{t \to i}(Z^{t}, Z^{i}) $), we use the softmax function to normalize these matrices in Eq. \ref{KL_divergence_similarity_matrix1} and Eq. \ref{KL_divergence_similarity_matrix2}. The supervisory matrix is normalized after matrix multiplication as in Eq. \ref{label_mask}. Similar to \cite{zhang2018deep}, since we project features from visual (or textual) modality into textual (or visual) modality, the KL-divergence regularizes the semantics in bi-directional feature projection, which is formulated in Eq. \ref{KL_divergence_symmetrical_loss} as:

\vspace{-1.0em}
\begin{equation}
\label{KL_divergence_similarity_matrix1}
\setlength\abovedisplayskip{3pt}
\setlength\belowdisplayskip{1pt}
\begin{aligned}
P_{i \to t} = \frac{exp(A_{i \to t}(Z^{i}, Z^{t}))}{\sum exp( A_{i \to t}(Z^{i}, Z^{t}))}
\end{aligned}
\end{equation}

\begin{equation}
\label{KL_divergence_similarity_matrix2}
\setlength\abovedisplayskip{3pt}
\setlength\belowdisplayskip{1pt}
\begin{aligned}
P_{t \to i} = \frac{exp(A_{t \to i}(Z^{t}, Z^{i}))}{\sum exp( A_{t \to i}(Z^{t}, Z^{i}))}
\end{aligned}
\end{equation}

\begin{equation}
\label{label_mask}
\setlength\abovedisplayskip{3pt}
\setlength\belowdisplayskip{1pt}
\begin{aligned}
Q_{y} = \frac{exp(Y_l^{\top}Y_l)}{\sum exp( Y_l^{\top}Y_l)}
\end{aligned}
\end{equation}

\begin{equation}
\label{KL_divergence_symmetrical_loss}
\setlength\abovedisplayskip{3pt}
\setlength\belowdisplayskip{1pt}
\begin{aligned}
L_{kl} & = L_{kl_{i \to t}} + L_{kl_{t \to i}} \\
     & = \frac{1}{N} \Big(\sum \sum P_{i \to t} \ast log \big( \frac{P_{i \to t}}{Q_{y} + \varepsilon}\big) 
      + \sum \sum P_{t \to i} \ast log\big( \frac{P_{t \to i}}{Q_{y} + \varepsilon}\big)\Big)
\end{aligned}
\end{equation}
where $\varepsilon $ is a small constant to avoid division by zero. Loss $L_{kl}$ refers to the KL-divergence between the projections of image-text features and their supervisory matrix. This loss is minimized and the gradients computed from $L_{kl}$ are used to update the parameters $\boldsymbol \theta_{E_{1}}$ and $\boldsymbol \theta_{E_{2}}$ of the generator, thereby the semantics between image features and text features can be associated.

\subsection{Instance Label Classification}
\label{Instance_Label_Classification}

\subsubsection{Categorical Cross-entropy Loss}
Label classification is a popular idea for cross-modal features learning \cite{zhang2018deep}. We use the instance labels provided on the datasets for label classification. For categorical cross-entropy loss, we apply the norm-softmax strategy and feature projection in \cite{zhang2018deep} to learn more discriminative cross-modal features. On the one hand, the normalized parameters $ \boldsymbol \theta_{P} $ in the label classifier encourage cross-modal features to distribute more compactly so that the softmax classifier performs label classification correctly. On the other hand, projection between image and text features strengthens their similarity association and is beneficial for label classification \cite{zhang2018deep}. Feature projection can be computed using Eq. \ref{Feature_projection}. Subsequently, given the instance label $ y_{l} $, categorical cross-entropy loss $ L_{ce} $ is defined by Eq. \ref{label_prediction}\footnote{We omit the bias term for simplicity} and is minimized during training:

\begin{equation}
\label{label_prediction}
\begin{aligned}
& L_{ce} = \mathbb{E}_{_{i,t}}(- y_{l} \ast log(p_{_{P}}(c|Z^{i,t};\boldsymbol \theta_{P}))) \\
= - \frac{1}{N} \Big( & \sum_{j=1}^{N} y_{l,j} \ast log\big(\frac{exp(\textbf{W}_{y_{l,j}}^{\top}\hat{z}^{i \to t}_{j})}{\sum_{j} exp(\textbf{W}_{j}^{\top}\hat{z}^{i \to t}_{j})}\big) + 
 \sum_{j=1}^{N} y_{l,j} \ast log\big(\frac{exp(\textbf{W}_{y_{l,j}}^{\top}\hat{z}^{t \to i}_{j})}{\sum_{j} exp(\textbf{W}_{j}^{\top}\hat{z}^{t \to i}_{j})}\big) \Big) \\
& s.t. \quad || \textbf{W}_{j} || = 1;  \hat{z}^{i \to t}_{j}= <\!\!z^{i}_{j} , \bar{z}^{t}_{j}\!\!> \ast \bar{z}^{t}_{j};
\hat{z}^{t \to i}_{j} = <\!\!z^{t}_{j} , \bar{z}^{i}_{j}\!\!> \ast \bar{z}^{i}_{j}
\end{aligned}
\end{equation}
where $ N $ is the number of image-text pairs in a mini-batch. $ W_{y_{l,j}} $ and $ W_{j} $ represent the $ y_{l,j} $-th and the $ j $-th column of weights $ \textbf{W} $ in classifier parameters $ \boldsymbol \theta_{P} $ according to \cite{zhang2018deep}. $ \hat{z}^{i \to t}_{j} $ and $ \hat{z}^{t \to i}_{j} $ are the projections image to text and the projections text to image, respectively, by using Eq. \ref{Feature_projection}.
 
 \subsubsection{KL-divergence for Data Imbalance}

Label classification using categorical cross-entropy loss can preserve semantic correlations between cross-modal features. However, we argue that there also exists a data imbalance issue when training the label classifier because each image is described by more than one sentence (\emph{e.g.} each image has five description sentences in the Flickr30K dataset). In the end, it causes the learned label classifier to prefer text features.
 
The issue of data imbalance in cross-modal retrieval can be resolved by constructing an augmented semantic space to re-align features \cite{zhong2020novel}. In this work, we use the temperature scaling \cite{guo2017calibration} to tackle the data imbalance issue. The biased label classifier can be calibrated by re-scaling its output probabilities \emph{i.e.}, $p^{i \to t} \!\! = \!\! softmax( \frac{\textbf{W}^{\top}\hat{z}^{i \to t}}{\tau} )  $ and $p^{t \to i} \! = \! softmax(\frac{\textbf{W}^{\top}\hat{z}^{t \to i}}{\tau})  $, respectively. Re-scaling the probabilities  with temperature $\tau$ raises the output entropy so better image-text matching can be observed \cite{guo2017calibration}. Subsequently, we use KL-divergence to measure the differences between the re-scaled probabilities. Since the magnitudes of the gradients produced by the re-scaling probabilities scale as $1/\tau^{2}$, it is important to multiply them by $\tau^{2}$. Finally, the KL-divergence loss on the scaling probabilities for data imbalance can be formulated as $ L_{di}$:

  \begin{equation}
 \label{KnowledgeDis}
\begin{aligned}
L_{di}   \! & =  \!\frac{\tau ^{2}}{N} \Big( \sum \sum   p^{i \to t} \! \ast \! log( \frac{p^{i \to t}}{p^{t \to i} \! + \! \varepsilon}) 
      + p^{t \to i} \! \ast \! log\big( \frac{p^{t \to i}}{p^{i \to t} \! + \! \varepsilon}\big) \Big) \\
   & s.t. \;\; p^{i \to t} \!\! = \!\! softmax\Big( \frac{\textbf{W}^{\top}\hat{z}^{i \to t}}{\tau} \Big), p^{t \to i} \! = \! softmax\Big(\frac{\textbf{W}^{\top}\hat{z}^{t \to i}}{\tau}\Big)  
\end{aligned}
\end{equation}
where $\varepsilon $ is a small constant to avoid division by zero. With $ \tau = 1 $, we recover the original KL-divergence. As reported in Table \ref{Distillation_temperature}, we find that the parameter $\tau$ can affect the effectiveness of loss $L_{di}$. Minimizing loss $L_{di}$ effectively reduces the influence of data imbalance issue and improves retrieval accuracy. The final objective function for label classification is ($ L_{ce} + L_{di} $). The gradients calculated from loss ($ L_{ce} + L_{di} $) are used to optimize the parameters $\boldsymbol \theta_{E_{1}}$, $\boldsymbol \theta_{E_{2}}$, and $ \boldsymbol \theta_{P} $ in the generator and the label classifier, respectively.

\subsection{Bi-directional Triplet Constraint}
\label{Bi_directional_Triplet_Constraint}

The triplet constraint is commonly used for feature learning. To achieve the baseline performance, we use this constraint from an inter-modality and an intra-modality perspective to strengthen the discrimination of cross-modal features. 

Given cross-modal features  $ Z^{i} $ and $  Z^{t}$ in the shared space, the cosine function is used to measure global similarity between feature vectors, \emph{i.e.} $ S_{jk} = (Z^{i}_{j})^{\top} Z^{t}_{k} $. We adopt the hard sampling strategy to select three-tuples features from an inter-modality and an intra-modality viewpoint. Hence, the inter-modality and intra-modality triplet loss functions are formulated as:

\begin{equation}
\setlength\abovedisplayskip{3pt}
\setlength\belowdisplayskip{1pt}
\begin{aligned}
L_{inter} = & \frac{1}{N}\Big(\!\sum_{j,k^{+},k^{-}}^{N}\!\!\!\!max[0, m-S_{j,k^{+}}+ S_{j,k^{-}}] 
			+   \sum_{k,j^{+},j^{-}}^{N}\!\!\!\!max[0, m-S_{k,j^{+}}+ S_{k,j^{-}}]\Big)
\end{aligned}
\end{equation}

\begin{equation}
\setlength\abovedisplayskip{3pt}
\setlength\belowdisplayskip{1pt}
\begin{aligned}
L_{intra} = & \frac{1}{N}\Big(\!\sum_{j,j^{+},j^{-}}^{N}\!\!\!\!max[0, m-S_{j,j^{+}}+ S_{j,j^{-}}] 
			+  \sum_{k,k^{+},k^{-}}^{N}\!\!\!\!max[0, m-S_{k,k^{+}}+ S_{k,k^{-}}]\Big)
\end{aligned}
\end{equation}

\begin{equation}
\label{Tripletloss}
\setlength\abovedisplayskip{3pt}
\setlength\belowdisplayskip{1pt}
\begin{aligned}
L_{tr} = L_{inter} + L_{intra}
\end{aligned}
\end{equation}
where $ m $ is the margin in the  bi-directional triplet loss function. For instance, in case of inter-modality, $ S_{j,k^{+}} = (Z^{i}_{j})^{\top} Z^{t}_{k^{+}}  $, where the anchor features are selected from the visual modality, while the positive features are selected from the textual modality. In case of intra-modality, $ S_{j,j^{+}} = (Z^{i}_{j})^{\top} Z^{i}_{j^{+}}  $, both the anchor features and the positive features are selected from the visual modality. Minimizing bi-directional triplet loss $L_{tr}$ keeps the correlated image-text pairs closer to each other, while the uncorrelated image-text pairs are pushed away. This loss directly operates on the cross-modal features  $ Z^{i} $ and $  Z^{t}$ so that the gradients from it optimize the parameters $\boldsymbol \theta_{E_{1}}$ and $\boldsymbol \theta_{E_{2}}$ of the generator.

\begin{algorithm}
\label{algorithm1}
 \caption{Whole network training and optimization pseudocode
}
 \begin{algorithmic}[1]
 \renewcommand{\algorithmicrequire}{\textbf{Input:}}
 \renewcommand{\algorithmicensure}{\textbf{Initialization:}}
 
 \Require mini-batch images $ X^{i} $, text $ X^{t} $, instance label $ Y $, modality label ($ Y^{i}_{c}$, $Y^{t}_{c}$), total training batch $ S $, pre-trained parameters $ \boldsymbol \theta_{E_{1}} $, update steps $ k$
 \Ensure  learning rate $ lr_{1} $, $ lr_{2} $, $ \boldsymbol \theta_{E_{2}} $, $ \boldsymbol \theta_{P} $, $ \boldsymbol \theta_{D} $
 \For { $ n =1$ to $ S $}
   \For {$k$ steps}
        \quad  \quad  \State cross-modal features embedding:
      	\State $ Z^{i} = E_{1}(X^{i}; \boldsymbol \theta_{E_{1}})$ \quad  \quad (Embed image features into the shared space)
   		\State $ Z^{t} = E_{2}(X^{t}; \boldsymbol \theta_{E_{2}})$ \quad  \quad (Embed text features into the shared space)
   		\quad  \quad  \State loss computing and feature optimization:
   		\State $ L_{ce}, L_{di}, L_{tr}, L_{kl}$ calculation   \quad \quad (Eqs. \ref{label_prediction}, \ref{KnowledgeDis}, \ref{Tripletloss}, \ref{KL_divergence_symmetrical_loss})
   		\State $ P^{i}_{D} = D(Z^{i}; \boldsymbol \theta_{D})$ \quad  \quad \quad(Discriminator $ D $)
   		\State $ P^{t}_{D} = D(Z^{t}; \boldsymbol \theta_{D})$
   		\State $ L_{s}, L_{c}$ calculation (Eqs. \ref{InformationEntropy_unified}, \ref{Modalityclassification_unified})
   		\quad  \quad  \State fix $ \boldsymbol \theta_{D} $, update parameters $ \boldsymbol \theta_{E_{1}} $, $ \boldsymbol \theta_{E_{2}} $, $ \boldsymbol \theta_{P} $:
   		\State $ \boldsymbol \theta_{P} \leftarrow \boldsymbol \theta_{P} - lr_{2} \cdot \nabla_{\boldsymbol \theta_{P}} (L_{ce}+L_{di}) $ 
   		\State $ \boldsymbol \theta_{E_{1}} \leftarrow \boldsymbol \theta_{E_{1}} - lr_{1} \cdot \nabla_{\boldsymbol \theta_{E_{1}}} (L_{ce}+L_{di}+ L_{tr}+ L_{kl}+ L_{s}) $ 
   		\State $ \boldsymbol \theta_{E_{2}} \leftarrow \boldsymbol \theta_{E_{2}} - lr_{2} \cdot \nabla_{\boldsymbol \theta_{E_{2}}} (L_{ce}+L_{di}+ L_{tr}+ L_{kl}+ L_{s}) $ 
   \EndFor
   \quad  \quad  \State fixate $ \boldsymbol \theta_{P} $, $ \boldsymbol \theta_{E_{1}} $, $ \boldsymbol \theta_{E_{2}} $, update parameters $ \boldsymbol \theta_{D} $:
   		\State $ \boldsymbol \theta_{D} \leftarrow \boldsymbol \theta_{D} - lr_{2} \cdot \nabla_{\boldsymbol \theta_{D}} ( L_{c}) $ 
   		
 \EndFor \\
 \Return the embedded cross-modal features $Z^{i}$ and $ Z^{t}$ in Figure \ref{ProblemSetting}
 \end{algorithmic} 
 \end{algorithm}
 
The problem of integrating information theory and adversarial learning for cross-modal retrieval is formally defined, in Eq. \ref{Opti}, as a min-max game using the previously defined loss terms. We further introduce the complete procedure of training and optimization in Algorithm 1. Finally, when trained to convergence, the network yields cross-modal features $Z^{i}$ and $ Z^{t}$ in the shared space, as shown in Figure \ref{ProblemSetting}. These return cross-modal features are used for performing retrieval.

\begin{equation}
\label{Opti}
\left\{
\begin{aligned}
 \mathop{min}\limits_{\boldsymbol \theta_{_{E_{1}}}, \boldsymbol \theta_{_{E_{2}}},\boldsymbol \theta_{_{P}}} & \mathop{max}\limits_{\boldsymbol \theta_{_{D}}}( L_{ce} + L_{di}  + L_{kl} + L_{tr} + L_{s} ) \\
 & \mathop{min}\limits_{\boldsymbol \theta_{_{D}}}L_{c}
 \end{aligned}
 \right.
\end{equation}

\section{Experiments}
\label{Experiments}

\subsection{Datasets and Settings}

We demonstrate the efficacy of the proposed method on the Flickr8K \cite{hodosh2013framing}, Flickr30K \cite{young2014image}, Microsoft COCO \cite{lin2014microsoft}, and CUHK-PEDES \cite{li2017person} datasets. Each image in these datasets is described by several descriptive sentences. For Flickr8K, we adopt the standard dataset splitting method to obtain a training set (6K), a validation set (1K), and a test set (1K). For Flickr30K, we follow the previous work \cite{zhang2018deep} and use 29,783 images for training, 1,000 images for validation and 1,000 images for testing. For  MS-COCO, we follow the training protocol in \cite{zhang2018deep} and split this dataset into 82,783 training, 30,504 validation and 5,000 test images, and then report the performance on both 5K and 1K test set. For CUHK-PEDES, it contains 40,206 pedestrian images of 13,003 identities. Following \cite{zhang2018deep}, we split this dataset into 11,003 training identities with 34,054 images, 1,000 validation identities with 3,078 images and 1,000 test identities with 3,074 images. Note that all captions for the same image are used as separate image-text pairs to train network.

Models are trained on GEFORCE TITAN X and Tesla K40 GPUs. To extract text features, the embedded words are fed into a Bi-LSTM to capture vectors with dimension 1024 (1024-D). We follow \cite{zhang2018deep} and set the Bi-LSTM with dropout rate 0.3. For fair comparison, we adopt ResNet \cite{he2016deep}, MobileNet \cite{howard2017mobilenets}, and VGGNet \cite{simonyan2014very} as the backbone to extract image features and further fine-tune them with learning rate $ lr_{1} \!= \! 2 \times 10^{-5} $, decaying every 2 epochs exponentially. The output 2048-D image features and 1024-D text features are further projected into a shared space. Then cross-modal features in the space are 512-D vectors (\emph{i.e.} $ Z^{i}$ and $ Z^{t} $ in Figure \ref{ProblemSetting}). The batch size is set to 64 or 32 depending on available GPUs memory. For the bi-directional triplet loss function, initially, we treat the inter-modality and intra-modality sampling identically although each of them might have different contributions \cite{wang2016learning}, we empirically set the margin to $ m = 0.5 $. The re-scaling parameter $\tau$ for data imbalance issue is set as $\tau=4$ (see Table \ref{Distillation_temperature}). In practice, the discriminator can classify image and text modality easily at the start of training, so the generator typically requires multiple (\emph{e.g.}, 5) update steps per discriminator update step during training (see Algorithm 1).

Once trained to converge, the network yields image features $Z^{i}$ and text features $ Z^{t}$. We use the cosine function to measure their similarity. We use Recall@K (K=1, 5, 10) for evaluation and comparison. Moreover, we adopt the precision-recall and mAP for the ablation studies, and visualize their feature distributions by t-SNE. Furthermore, we display the cross-modal retrieval results using our method.

\begin{table*}[!t] 
\footnotesize
\vspace{-1.5em}
\centering 
\caption{Comparison of retrieval results on the Flickr30K \cite{young2014image} and MS-COCO \cite{lin2014microsoft} dataset (R@K (K=1,5,10)(\%))}
\vspace{-1.0em}
\setlength{\tabcolsep}{1mm}
\begin{tabular}{p{82 pt}c|c|c|c|c|c|c|c|c|c|c|c|c} 
\hline 

\multicolumn{2}{c|}{ \multirow{2}*{} } & \multicolumn{6}{c|}{Flickr30K} & \multicolumn{6}{c}{MS-COCO} \\ 
\cline{1-14} 

\multicolumn{2}{c}{ \multirow{2}*{ Method } } \multirow{2}*{Backbone Net}  & \multicolumn{3}{|c|}{Image-to-Text} & \multicolumn{3}{c}{Text-to-Image} & \multicolumn{3}{|c|}{Image-to-Text} & \multicolumn{3}{c}{Text-to-Image}\\ 
\cline{3-14} 
\multicolumn{1}{c}{} & & R@1 & R@5 & R@10 & R@1 & R@5 & R@10 & R@1 & R@5 & R@10 & R@1 & R@5 & R@10 \\ 
\hline 
 m-RNN \cite{mao2014deep} & VGG & 35.4 & 63.8 & 73.7 & 22.8 & 50.7 & 63.1 & 41.0 & 73.0 & 83.5 & 29.0 & 42.2 & 77.0\\ 
 
  RNN+FV \cite{lev2016rnn} & VGG & 35.6 & 62.5 & 74.2 & 27.4 & 55.9 & 70.0 & 41.5 & 72.0 & 82.9 & 29.2 & 64.7 & 80.4 \\ 

  DSPE+FV \cite{wang2016learning} & VGG & 40.3 & 68.9 & 79.9 & 29.7 & 60.1 & 72.1 & 50.1 & 79.7 & 89.2 & 39.6 & 75.2 & 86.9\\ 

  CMPM+CMPC$^{\dagger}$ \cite{zhang2018deep}  & MobileNet &  40.3 & 66.9 & 76.7 & 30.4 & 58.2 & 68.5 & 52.9 & 83.8 & 92.1 & 41.3 & 74.6 & 85.9\\
  
Word2VisualVec \cite{dong2018predicting}  & ResNet-152 & 42.0 & 70.4 & 80.1 & - & - & - & - & - & - & - & - & - \\

sm-LSTM \cite{huang2017instance} & VGG & 42.5 & 71.9 & 81.5 & 30.2 & 60.4 & 72.3 & 53.2 & 83.1 & 91.5 & 40.7 & 75.8 & 87.4\\ 

 RRF-Net \cite{liu2017learning}  & ResNet-152  & 47.6 & 77.4 & 87.1 & 35.4 & 68.3 & 79.9 & 56.4 & 85.3 & 91.5 & 43.9 & 78.1 & 88.6\\

Joint learning \cite{wang2019cross} & ResNet-152 & 48.6 & 73.6 & 83.6 & 32.3 & 62.5 & 74.0 & 55.3 & 82.7 & 90.2 & 41.7 & 75.0 & 87.4\\

 CMPM+CMPC$^{\ddagger}$  \cite{zhang2018deep} &  ResNet-152 & 49.6 & 76.8 & 86.1 & 37.3 & 65.7 & 75.5 & - & - & - & - & - & - \\
 
 
 VSE++ \cite{faghri2017vse}  & ResNet-152 & 52.9 & 80.5 & 87.2 & 39.6 & 70.1 & 79.5 & 51.3 & 82.2 & 91.0 & 40.1 & 75.3 & 86.1\\
 
  TIMAM  \cite{sarafianos2019adversarial}  & ResNet-152 & 53.1 & 78.8 & 87.6 & 42.6 & 71.6 & 81.9 & - & - & - & - & - & - \\

  DAN \cite{nam2017dual}  & ResNet-152 & 55.0 & 81.8 & 89.0 & 39.4 & 69.2 & 79.1 & - & - & - & - & - & -\\
  
  Dual-path stage I \cite{zheng2017dual} & ResNet-152 & 44.2 & 70.2 & 79.7 & 30.7 & 59.2 & 70.8 & 52.2 & 80.4 & 88.7 & 37.2 & 69.5 & 80.6 \\
  
  Dual-path stage II \cite{zheng2017dual} & ResNet-152 & 55.6 & 81.9 & 89.5 & 39.1 & 69.2 & \textbf{80.9} & \textbf{65.6} & \textbf{89.8} & \textbf{95.5} & \underline{47.1} & 79.9 & \underline{90.0} \\
\hline 

Our ITMeetsAL& VGG & 38.5 & 66.5 & 76.3 & 30.7 & 59.4 & 70.3 & 44.2 & 76.1 & 86.3 & 37.1 & 72.7 & 85.1\\
 
Our ITMeetsAL & MobileNet & 46.6 & 73.5 & 82.5 & 34.4 & 63.3 & 74.2 & 54.7 & 84.3 & 91.1 & 41.0 & 76.7 & 88.1\\

Our ITMeetsAL & ResNet-152 &  \textbf{56.5} & \textbf{82.2} & \textbf{89.6} & \textbf{43.5} & \textbf{71.8} & 80.2  & \underline{58.5} & 85.3 & \underline{92.1} & \textbf{48.3} & \textbf{82.0} & \textbf{90.6} \\
\hline
\end{tabular} 
\footnotesize{ MS-COCO is tested on 1K images. The best results are in bold and the second best results are underlined. }
\end{table*}

\subsection{Performance Evaluation}

\subsubsection{Results on the Flickr30K and MS-COCO Datasets}

The retrieval results on the Flickr30K and MS-COCO datasets are reported in Table 1. Hereafter, ``Image-to-Text'' means using an image as a query item to retrieve semantically-relevant text from the textual gallery. ``Text-to-Image'' means using a text as query to retrieve images from the visual gallery. In most cases, our proposed approach shows the best performance when using three different deep networks. For the ``Image-to-Text'' task on the MS-COCO dataset, the best results are obtained by Zheng \etal \cite{zheng2017dual}, which adopted a deeper network for text feature learning and used a two-stage training strategy. However, for the ``Text-to-Image'' task and the ``Image-to-Text'' task on the Flickr30K dataset, our method performs better. Take ResNet-152 as an example, the results are R@1=43.5\% on the Flickr30K and R@1=48.3\% on the MS-COCO for ``Text-to-Image'' task; the results are R@1=56.5\% on the Flickr30K  dataset and R@1=58.5\% on the MS-COCO dataset for ``Image-to-Text'' task.

Besides, we obverse that the strategy for network training is critical for retrieval performance. Take \cite{zheng2017dual} as an example, the backbone network (ResNet-152) is fixed at stage I ( R@1=44.2\% on ``Image-to-Text'' task on Flickr30K) and then fine-tuned with a small learning rate on stage II (R@1=55.6\% on the ``Image-to-Text'' task on Flickr30K). In contrast, our network structure is trained end-to-end in only one stage (we fine-tune the backbone network with a small learning rate from the beginning). Our reported results are close to those in two-stage dual learning \cite{zheng2017dual}. When tested on the Flickr30K dataset for the ``Image-to-Text'' task, the recall results are R@1=56.5\%, R@5=82.2\%, R@10=89.6\%, which are the best overall previous methods.

Obviously, the feature learning capacity of the backbone networks would affect retrieval performance significantly. We can see from Table 1, the retrieval results based on ResNet-152 are usually higher than those of MobileNet and VGGNet. Moreover, our method also has good performance using MobileNet. For instance, regarding the ``Image-to-Text'' task on the Flickr30K dataset, the recall result of CMPM+CMPC \cite{zhang2018deep} is R@1=40.3\%, but the result from our method is R@1=46.6\%, which is a significant improvement.

Considering the two branches of ``Image-to-Text'' task and the ``Text-to-Image'' task, we think that the data imbalance issue still influences the performance of each branch. More specifically, for all listed methods, the ``Image-to-Text'' task has better performance, which indicates that the network still has more biases on text feature learning as a result of the issue of data imbalance. Thus, there exists more room for improvement using other strategies, such as data augmentation.

\begin{table}[!t] 
\footnotesize
\vspace{-1.5em}
\centering 
\caption{Retrieval results on the CUHK-PEDES \cite{li2017person} dataset (R@K (K=1,5,10)(\%))}
\vspace{-1.0em}
\setlength{\tabcolsep}{3mm}
\begin{tabular}{p{103 pt}c|c|c|c} 
\hline 

\multicolumn{2}{c}{ \multirow{2}*{ Method} } \multirow{2}*{Backbone Net} &  \multicolumn{3}{|c}{Text-to-Image} \\ 
\cline{3-5} 
\multicolumn{1}{c}{} & & R@1 & R@5 & R@10 \\ 
\hline 
Latent co-attention \cite{li2017identity} & VGG & 25.94 & - & 60.48\\ 

Local-global association \cite{chen2018improving} & ResNet-50 & 43.58 & 66.93 & 76.26 \\

CMPM \cite{zhang2018deep} & MobileNet & 44.02 & - & 77.00\\ 

Dual-path two-stage \cite{zheng2017dual} & ResNet-152 & 44.40 & 66.26 & 75.07 \\ 

MIA \cite{niu2019improving} &  ResNet-50 & 48.00 & 70.70 & 79.30\\ 

CMPM+CMPC \cite{zhang2018deep} & MobileNet & 49.37 & - & 79.27\\ 
 
\hline 
 
  Our ITMeetsAL  & VGG &  44.43 & 68.26 & 77.50 \\
  
 Our ITMeetsAL  & MobileNet &  51.85 & 73.36 & 81.27 \\

 Our ITMeetsAL  & ResNet-50 & 50.63 & 73.33 & 81.34  \\
 
  Our ITMeetsAL  & ResNet-152 & \textbf{55.72} & \textbf{76.15} & \textbf{84.26}  \\
\hline
\end{tabular} 
\end{table}

\subsubsection{Results on CUHK-PEDES Dataset}

The ``Text-to-Image'' retrieval results on the CUHK-PEDES dataset are reported in Table 2. We evaluate the proposed method using four deep networks. All results indicate that our method outperforms other counterparts. The optimal results are achieved with R@1=55.72\% using ResNet-152 as backbone network. The results using MobileNet are sub-optimal but also have some improvements. For example, CMPM+CMPC achieves a recall R@1=49.37\% and R@10=79.27\%, while our method obtains R@1=51.85\% and R@10=81.27\%. Moreover, the results of our method show that deeper networks achieve better retrieval performance, whereas the light-weight MobileNet has a similar performance as ResNet-50.

\subsubsection{Results on Flickr8K Dataset}

The retrieval results on the Flick8K dataset are reported in Table 3. The best results R@1=40.6\%, R@5=67.8\%, R@10=78.6\% are achieved by joint correlation learning \cite{wang2019cross} where a batch-based triplet loss, which considers all image-sentences pairs, is used for learning correlations. The second-best results are achieved using ResNet-152 (same as \cite{wang2019cross}) R@1=40.1\%, R@5=67.8\%, R@10=79.2\%, which has better R@10 performance compared to \cite{wang2019cross}. Our method shows competitive results compared to other counterparts and also indicates that there exists room for further performance improvement.

\begin{table}[!t] 
\footnotesize
\vspace{-1.5em}
\centering 
\caption{Retrieval results on the Flickr8K \cite{hodosh2013framing} dataset (R@K (K=1,5,10)(\%))}
\vspace{-1.0em}
\setlength{\tabcolsep}{3mm}
\begin{tabular}{p{78 pt}c|c|c|c} 
\hline 

\multicolumn{2}{c}{ \multirow{2}*{ Method} } \multirow{2}*{Backbone Net} &  \multicolumn{3}{|c}{Image-to-Text} \\ 
\cline{3-5} 
\multicolumn{1}{c}{} & & R@1 & R@5 & R@10 \\ 
\hline 

RNN+FV \cite{lev2016rnn} & VGG & 23.2 & 53.3 & 67.8 \\

GMM+HGLMM \cite{klein2015associating} & VGG & 31.0 & 59.3 & 73.7\\ 

Word2VisualVec \cite{dong2018predicting} & ResNet-152 & 33.4 & 63.1 & 75.3 \\

Joint learning \cite{wang2019cross} & ResNet-152 & \textbf{40.6} & \textbf{67.8} & \underline{78.6}\\ 
\hline 

Our ITMeetsAL  & VGG & 28.0 & 52.7 & 63.1 \\

 Our ITMeetsAL  & MobileNet & 30.9 & 58.6 & 70.8 \\

  Our ITMeetsAL  & ResNet-152 & \underline{40.1} & \textbf{67.8} & \textbf{79.2} \\
\hline
\end{tabular} 
\\
\footnotesize{The best results are in bold and the second best results are underlined. }
\end{table}

\subsection{Ablation Studies}

For analyzing the effect of each component, the ablation studies are conducted on the Flickr30K dataset using MobileNet as a backbone net, we use the commonly used categorical cross-entropy $L_{ce}$ and bi-triplet loss function $L_{tr} $ to construct the baseline in Table 4, we call this \textbf{Baseline1} configuration ``Only $ L_{ce}+ L_{tr}$''. 

\begin{table}[!t] 
\footnotesize
\centering 
\caption{ Component analysis on the Flickr30K \cite{young2014image} (R@1, R@10, and mAP (\%))}
\vspace{-1.0em}
\setlength{\tabcolsep}{1.0mm}
\begin{tabular}{p{55 mm}|c|c|c|c|c|c} 
\hline 

\multicolumn{1}{c|}{ \multirow{2}*{} } & \multicolumn{6}{c}{Flickr30K}  \\ 
\cline{2-7} 

\multicolumn{1}{c|}{ \multirow{2}*{ Method using MobileNet} } & \multicolumn{3}{c|}{Image-to-Text} & \multicolumn{3}{c}{Text-to-Image} \\
\cline{2-7} 
\multicolumn{1}{c|}{}& R@1 & R@10 & mAP & R@1 & R@10 & mAP \\ 
\hline 

Baseline1: Only $L_{ce} $+$L_{tr} $  & 40.6 & 80.8 & 23.1 & 31.9 & 72.2 & 31.9 \\

 Baseline2: $L_{ce} $+$L_{tr} $+$L_{di} $ & 42.3 & 80.6 & 24.4 & 32.5 & 73.0 & 32.5 \\

 Baseline3: $L_{ce} $+$L_{tr} $+$L_{di} $+$L_{kl} $  & 44.7 & 81.0 & 25.2 & 32.6 & 73.2 & 32.6 \\
  
Full method: $L_{ce} $+$L_{tr} $+$L_{di} $+$L_{kl} $+$L_{s}$+$L_{c}$ & 46.6 & 82.5 & 26.3 & 34.4 & 74.1 & 34.4 \\
 
\hline
\end{tabular}
\end{table}

\subsubsection{Analysis of KL-divergence for Data Imbalance}

Each image in a dataset (\emph{e.g.} Flickr30k) has more than one description sentence. We think this leads to a data imbalance issue for cross-modal feature learning. The network has more text data for training, which causes the learned label classifier to prefer text features. Therefore, we adopt a regularization term $L_{di} $ based on KL-divergence to calibrate this bias. To this end, the label classifier can be re-calibrated on the image features and text features. In Table 4, this \textbf{Baseline2} configuration is named `` $L_{ce}+L_{tr} +L_{di} $''. The Recall and mean Average Precision (mAP) show the effectiveness of this loss. Compared to Baseline1, the scaling KL-divergence loss $L_{di} $ contributes more on Recall@1 for both the ``Image-to-Text'' (42.3\%) and ``Text-to-Image'' task (32.5\%).

\subsubsection{Analysis of KL divergence for Cross-modal Feature Projection}

KL divergence is obtained by adding $L_{kl} $ which constrains the image features and text features in the shared space under the supervision of supervisory matrix. It focuses on the whole feature distribution and is complementary to the bi-directional triplet loss function. We denote \textbf{Baseline3} as ``$ L_{ce}+L_{tr}+L_{di}+L_{kl} $'' in Table 4. As we can see, Recall@1 of the ``Image-to-Text'' task has been improved significantly by 2.4\%. However, the KL-divergence loss shows a slight improvement on the ``Text-to-Image'' task. The results indicate that the KL-divergence loss function contributes more to image feature learning, which might be caused by the issue of data imbalance of the dataset.

\subsubsection{Analysis of Adversary Combining}

The prior loss terms have been used to constrain the similarity of the image-text features in the shared space. Intuitively, two-tuple or three-tuple feature exemplars are helpful for reducing the ``semantic gap'' and further making the whole feature distribution close at the same time. However, the constraint loss functions (\emph{e.g.} cosine similarity) cannot constrain the distribution discrepancy of the whole distribution because these loss functions are symmetrical. Focusing on the whole feature distribution, we combine the Shanon information entropy $ L_{s}$ and the modality classification loss $L_{c}$ in an adversary training manner to reduce the heterogeneity gap. This \textbf{full method} is named ``$ L_{ce}+L_{tr}+L_{di}+L_{kl}+L_{s}+L_{c} $'' and corresponding results are shown in Table 4. Compared to former baselines, the results obtained by using our method are improved significantly. 

\begin{figure}
\centering  
  \subfigure[]
 { \label{MetricLearning_a}     
   \includegraphics[width=0.4\columnwidth]{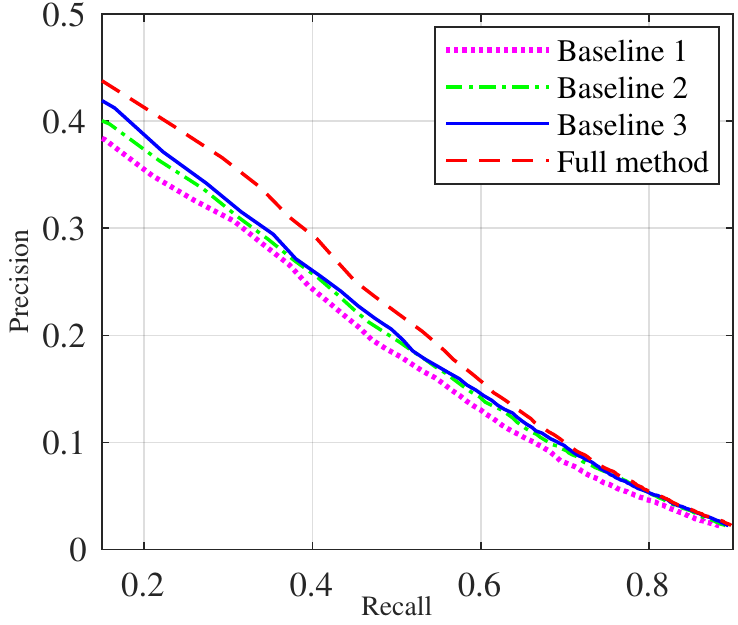}  
 }     
  \subfigure[] 
 { \label{MetricLearning_b}   
   \includegraphics[width=0.41\columnwidth]{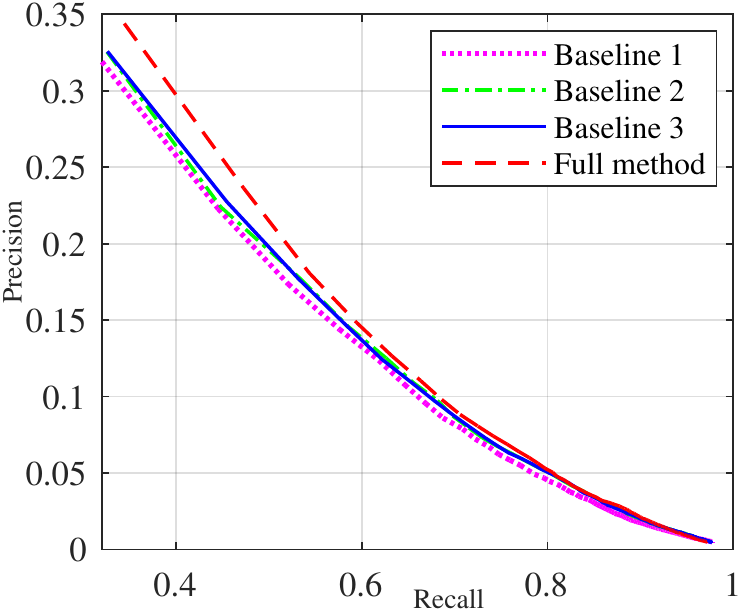}   
  }  
\vspace{-1.0em}
  \caption{ The precision\_recall curves from ``Baseline1'' to ``Full method'' on Flickr30K, each line corresponds one experimental configuration in Table 4. The larger area under the line indicates better performance.}  
   \label{Precision_recall_curve}   
 \end{figure}

Furthermore, we compare the precision-recall curves for the above four configurations and baselines, the results are shown in Figure \ref{Precision_recall_curve}. The larger the area under the curve, the better the algorithm. Regarding the different tasks, the improvements are slightly different. Overall, we can see  that each added component helps to improve the overall performance of the retrieval algorithm.

 \subsubsection{Analysis of Temperature $\tau$}
 
We analyze the temperature parameter $\tau$ in loss $L_{di}$ in Eq. \ref{KnowledgeDis}. Other loss terms are kept the same with the full method, \emph{i.e.} ``$ L_{ce}+L_{tr}+L_{di}+L_{kl}+L_{s}+L_{c} $''. We vary this parameter $\tau$ from 1 to 6, and their corresponding results are reported in Table \ref{Distillation_temperature}. We can observe that the optimal results are achieved if the classifier's output probabilities are re-scaled by $\tau=4$. As claimed in \cite{guo2017calibration}, the temperature scaling raises the output entropy of the classifier with $\tau>1$. In our experiments, we found it is beneficial for improving the image-text matching.

\begin{table}[!t] 
\footnotesize
\vspace{0em}
\centering 
\caption{ Temperature scaling analysis for loss $L_{di}$ (R@1, R@10, and mAP (\%))}
\label{Distillation_temperature}
\vspace{-1.0em}
\setlength{\tabcolsep}{1.0mm}
\begin{tabular}{c|c|c|c|c|c|c} 
\hline 

\multicolumn{1}{c|}{ \multirow{2}*{} } & \multicolumn{6}{c}{Flickr30K}  \\ 
\cline{2-7} 

\multicolumn{1}{c|}{ \multirow{1}*{Temperature} } & \multicolumn{3}{c|}{Image-to-Text} & \multicolumn{3}{c}{Text-to-Image} \\
\cline{2-7} 
\multicolumn{1}{c|}{}& R@1 & R@10 & mAP & R@1 & R@10 & mAP \\ 
\hline 

$ \tau $=1  & 44.0 & 80.6 & 24.8 & 32.9  & 73.5 & 32.9 \\
$ \tau $=2  & 45.3 & 80.9 & 25.6 & 33.6  & 73.6 & 33.6 \\
$ \tau $=3 & 46.2 & \textbf{83.2} & 25.7 & 33.3  & 73.4 & 33.3\\
$ \tau $=4 & \textbf{46.6} & 82.5 & \textbf{26.3} & \textbf{34.4} & \textbf{74.2}  & \textbf{34.4}\\
$ \tau $=5 & 46.0  & 81.6 & 26.1 & 34.3  & 73.9 & 34.3\\
$ \tau $=6 & 45.9 & 80.2 & 26.1 & 33.1 & 73.4 & 33.1  \\
\hline
\end{tabular}
\end{table}

\subsubsection{Distribution Visualization}

\begin{figure}[!t]
\centering  
  \subfigure[]
 { \label{Baseline1}     
   \includegraphics[width=0.4\columnwidth]{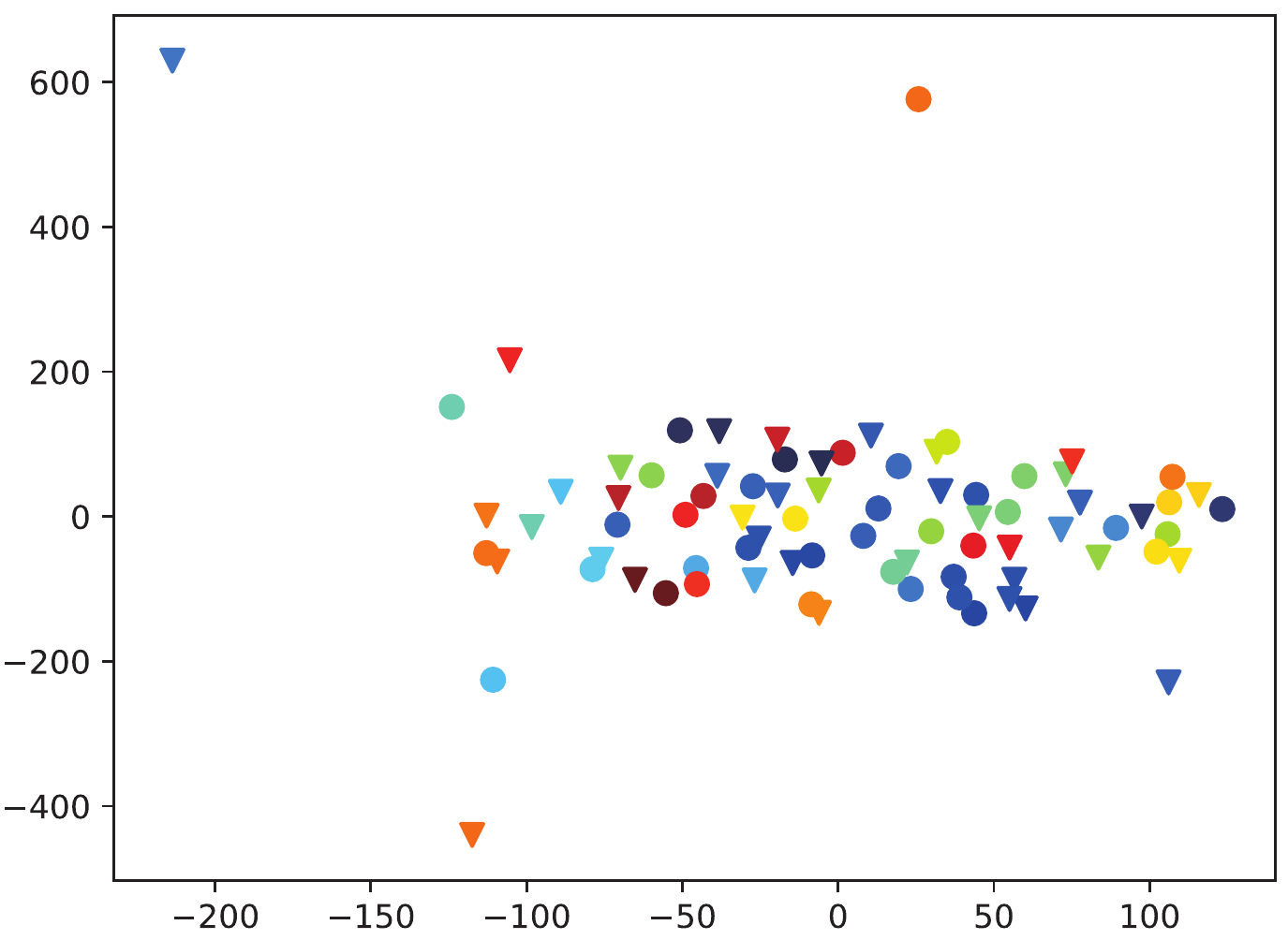} 
 }     
  \subfigure[] 
 { \label{Without_KL}   
   \includegraphics[width=0.4\columnwidth]{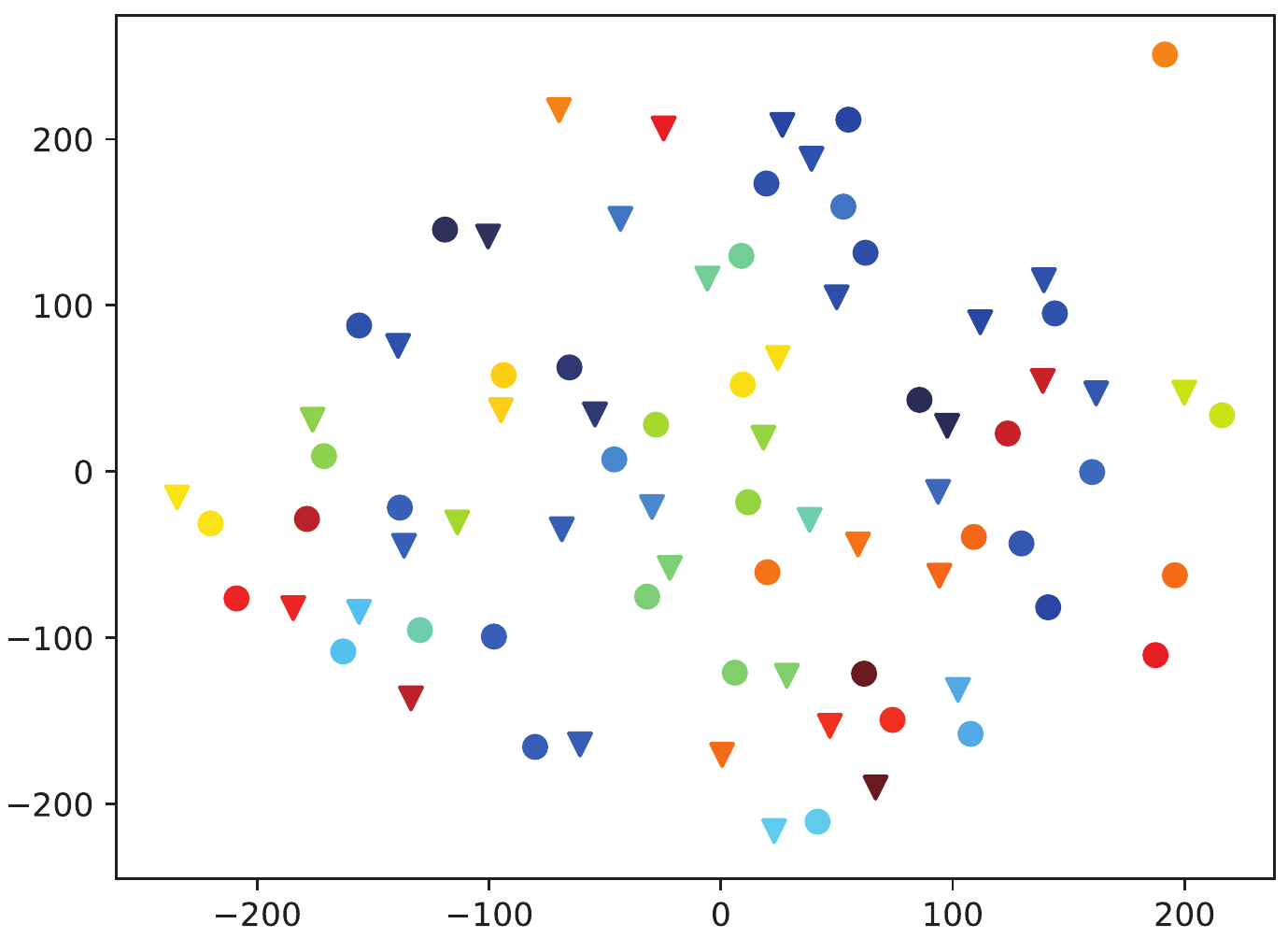}   
  }  

 \subfigure[]
 { \label{without_adversary}     
   \includegraphics[width=0.417\columnwidth]{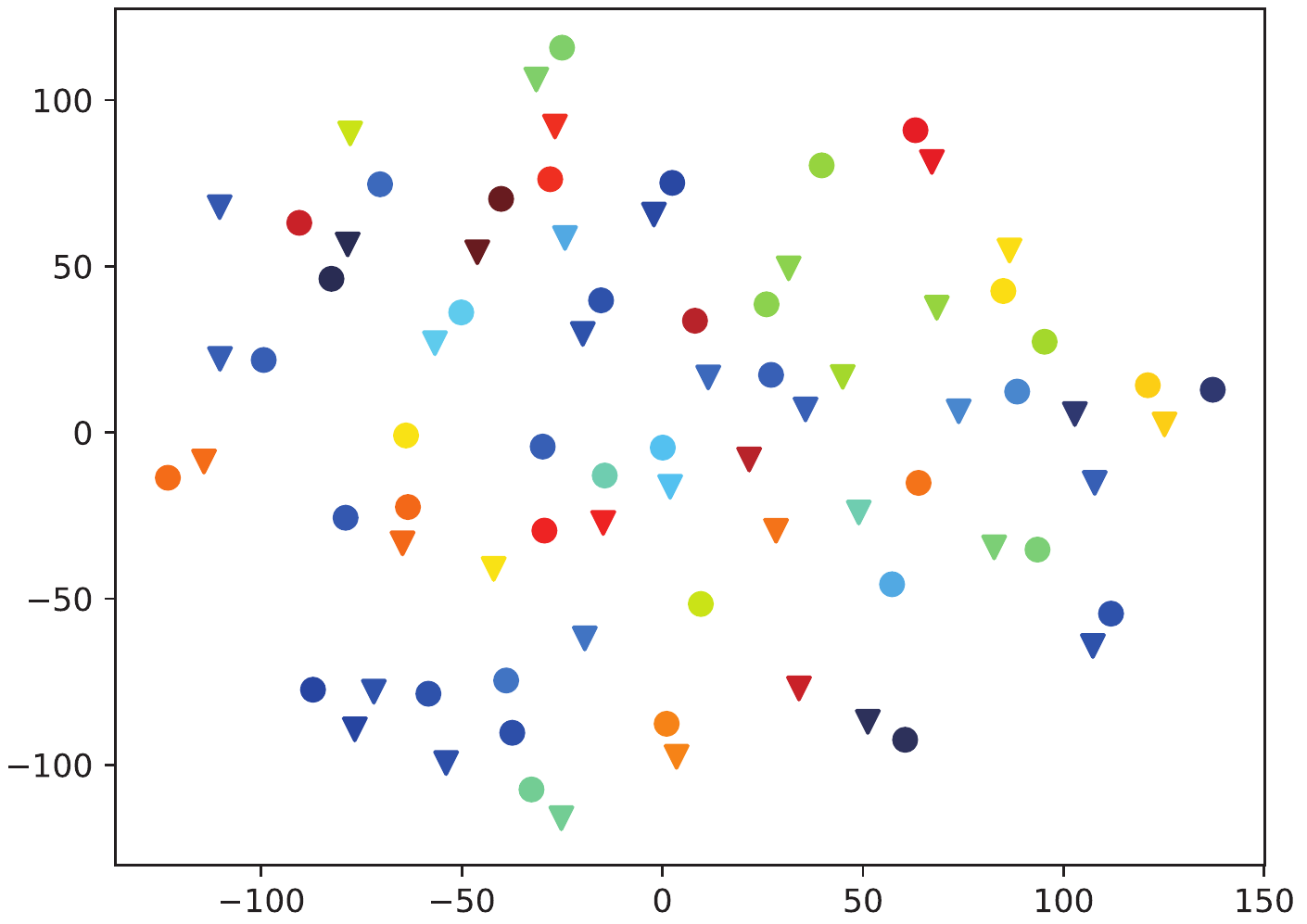}  
 }     
  \subfigure[] 
 { \label{Full_method_choosed}   
   \includegraphics[width=0.4\columnwidth]{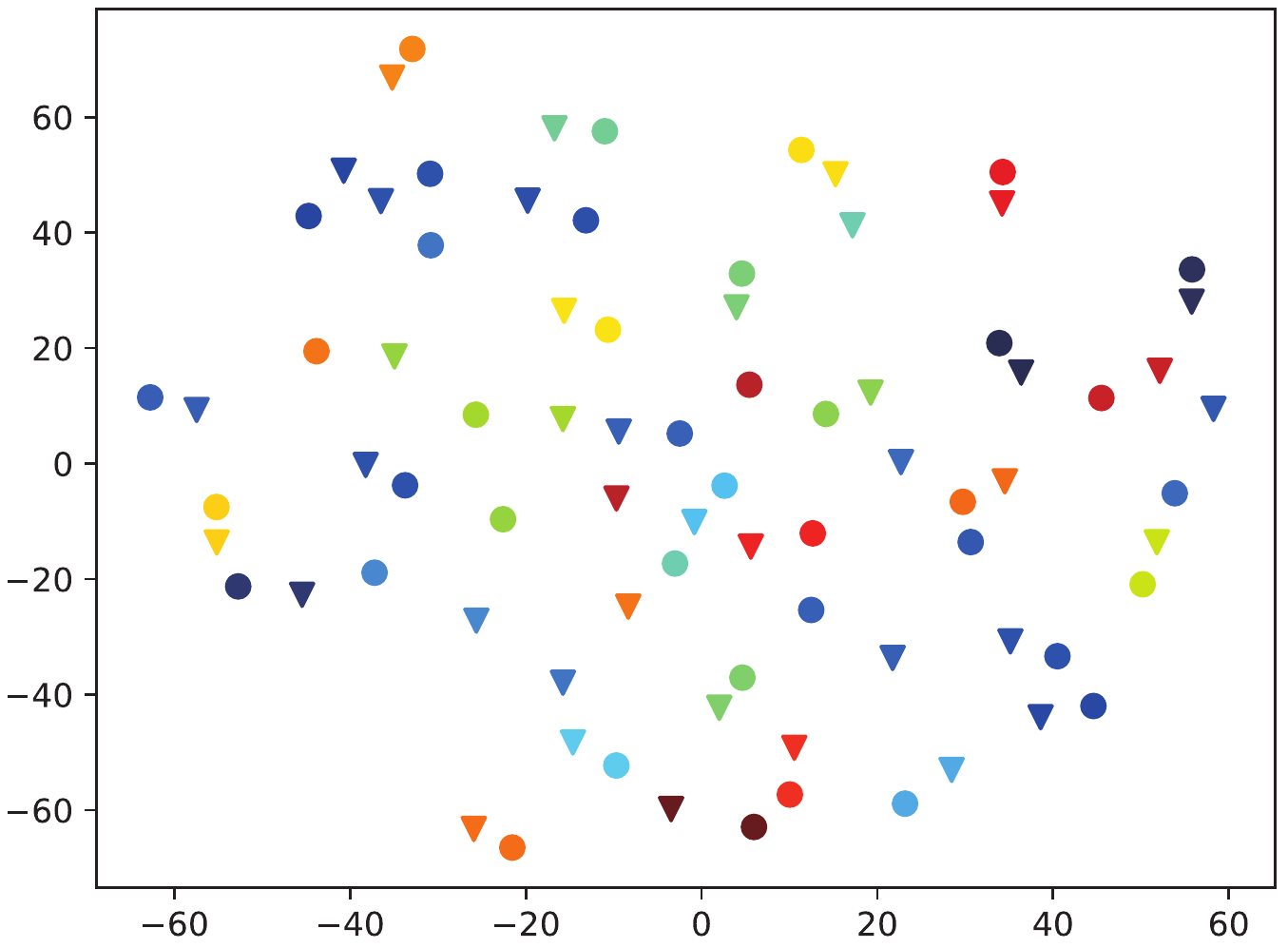}   
  }
  \caption{Feature distribution visualizations for the ablation studies. The shape represents modality and the color indicates the label information. Sub-figures (a)$ \sim  $(d) correspond to the four experimental configurations in Table 4. When each loss function is gradually applied, the paired image features and text features have smaller distances. Best viewed in color.}  
   \label{tSNE}   
 \end{figure}
 
We choose 40 image-text pairs from the Flickr30K dataset to visualize their feature distributions using t-SNE. We only choose the first description caption among the five sentences. In Figure \ref{tSNE}, the circle and the triangle shape denote text features and image features, respectively. Label information is represented by a different color.

This distribution indicates the effectiveness of each component (\emph{e.g.} KL-divergence for cross-modal feature projection, and the Shannon information entropy trained in an adversarial manner). In Figure \ref{Baseline1}, there exist several feature outliers within the distribution and the proximity relationship between pair-wise features is not obvious. When using the proposed components, the features distribute much better. For example, in Figure \ref{Full_method_choosed}, all loss functions are utilized to constrain feature learning, the pair-wise feature shows a close proximity relationship. Moreover, image features and text features are distributed within smaller ranges (-60 $ \sim $ 60). Few outliers exist among the whole distribution.

Qualitative retrieval results on the Flickr30K and the CUHK-PEDES dataset are shown in Figure \ref{Result_Retrieval_show}. For the ``Image-to-Text'' task, the proposed method can return almost all paired text of the query image. The ``Image-to-Text'' task also has good performance, the proposed method retrieves the paired image correctly. Also, other retrieved images show contents relevant to the query sentence.

\begin {figure}[!t]
\centering
 { 
   \includegraphics[width=0.6\columnwidth]{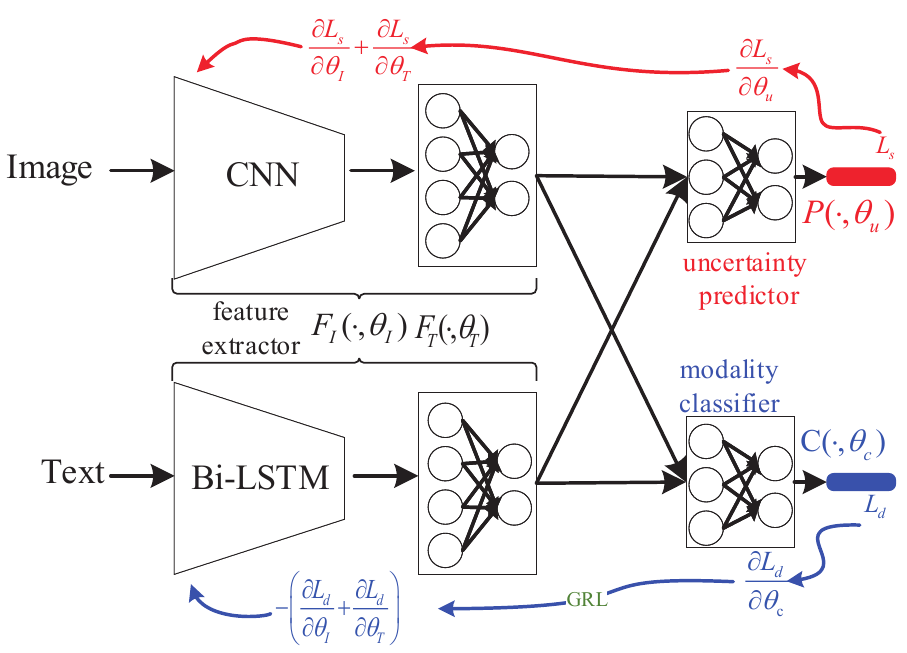} 
 }
\caption{ The illustration of independent combining information entropy and modality classification into an adversary, which is an intuitive structure of the diagram in Figure \ref{ProblemSetting}. Other loss functions including categorical cross entropy loss, KL-divergence loss, and bi-directional triplet loss are kept the same, but we do not show in this graph for simplicity. Different from the framework in Figure \ref{UnifiedDiscriminator}, the gradients computed from the modality classifier in this combining paradigm are used to optimize the parameters $ \boldsymbol \theta_{I} $ and $ \boldsymbol \theta_{T} $ of the feature extractor. The feature extractor \textit{maximizes} the loss $L_d = L_c$ (Eq. \ref{Modalityclassification_unified}) of modality classifier $C$ (to make image features and text features as similar as possible), while the parameters $ \boldsymbol \theta_{c} $ of the modality classifier \textit{minimize} the
loss $L_d$. This process depends on a gradient reversal layer to multiply gradient values by -1 when back-propagating \cite{ganin2015unsupervised}.} 
\label{SeparatedFramework}
\end {figure}

\subsection{Further Exploring}

In this paper, we propose to integrate Shannon information entropy with the discriminator for cross-modal retrieval. That is, the discriminator performs modality classification and measures the information entropy at the same time (see Figure \ref{UnifiedDiscriminator}). Herein, we further explore a paradigm to integrate information entropy with adversarial learning. This combining paradigm is more straightforward to the structure in Figure \ref{ProblemSetting}. Concretely, we build two branches of sub-networks: an uncertainty predictor for modality uncertainty prediction and a modality classifier for modality classification. Then adversarial learning is implemented as an interplay between these two-subnetworks with competitive objectives. The uncertainty predictor aims at maximizing the modality uncertainty of the shared space (measured by information entropy), while the modality classifier is to identify image inputs and text inputs by modality classification. We illustrate this combining paradigm in Figure \ref{SeparatedFramework}. Compared to the former paradigm depicted in Figure \ref{UnifiedDiscriminator}, the optimization depicted in Figure \ref{SeparatedFramework} is different and more complex. The gradients computed by the classifier are used to update parameters $ \boldsymbol \theta_{I} $ and $ \boldsymbol \theta_{T} $ in the feature extractor. To learn modality-invariant features, the feature extractor \textit{minimizes} the loss of the uncertainty predictor and it \textit{maximizes} the loss $L_d$ of the modality classifier, which aims to make image features and text features as similar as possible \cite{ganin2015unsupervised}. The parameters of the modality classifier \textit{minimize} its
loss $L_d$. This training process needs to depend on the gradient reversal layer \cite{ganin2015unsupervised}, which would multiply gradient values by -1 when executing back-propagating.

The training procedure is almost the same as used in Algorithm 1 except for the gradients from the modality classification loss that updates the backbone network, leading to a slower training process. The retrieval performance of these two combined methods presented in Figure \ref{UnifiedDiscriminator} and Figure \ref{SeparatedFramework} (named as unified and separate, respectively) are given in Table 5. The backbone net for image feature extraction is ResNet-152. These two combined strategies show different performances on the four datasets when combining information entropy and modality classification into a unified discriminator. The performance improves slightly on the Flickr30K, MS-COCO, and Flickr8K datasets when adopting the combining strategy of Figure \ref{UnifiedDiscriminator}. However, the method depicted in Figure \ref{SeparatedFramework} has better performance on the CUHK-PEDES dataset, which is not the common objects dataset. This method has R@1 improved by 3.3\% (from 65.58\% to 67.79\%), Also, the mAP has improved by 1.8\% compared to the unified method depicted in Figure \ref{UnifiedDiscriminator}. In summary, the proposed framework of combining information entropy and adversarial learning in Figure \ref{UnifiedDiscriminator} has better performance and has faster convergence during training.

 \begin{table*}[!t] 
 \footnotesize
\vspace{-1.5em}
\centering 
\caption{ Comparison of two combining paradigms in four retrieval datasets (R@1, R@10, and mAP(\%))}
\vspace{-1.0em}
\setlength{\tabcolsep}{0.7mm}
\begin{tabular}{p{76 pt}c|c|c|c|c|c|c|c|c|c|c|c|c} 
\hline 

\multicolumn{1}{c}{ \multirow{2}*{} } & \multicolumn{12}{c}{Image-to-Text} \\
\cline{1-14} 

\multicolumn{2}{c}{ \multirow{2}*{ Combining strategy} \multirow{2}*{ Backbone Net}} & \multicolumn{3}{|c|}{Flickr30K} & \multicolumn{3}{c}{MS-COCO} & \multicolumn{3}{|c|}{CUHK-PEDES} & \multicolumn{3}{c}{Flickr8K}\\ 
\cline{3-14} 
\multicolumn{1}{c}{} & & R@1 & R@10 & mAP & R@1 & R@10 & mAP & R@1 & R@10 & mAP & R@1 & R@10 & mAP \\ 
\hline 
 
Method in Figure \ref{SeparatedFramework} & ResNet-152 & 55.30 & 88.30 & 32.23 & 57.00 & 92.10 & 35.12 & 67.79 & 93.75 & 34.79 & 39.00 & 77.70 & 22.33\\

Method in Figure \ref{UnifiedDiscriminator} & ResNet-152 & 56.50 & 89.60 & 32.58 & 58.50 & 92.10 & 36.28 & 65.58  & 93.60 & 34.17 & 39.90 & 77.90 & 22.46\\
\hline
\end{tabular} 
\end{table*}

\begin{figure*}[t]
\centering
\includegraphics[width=1.0\textwidth]{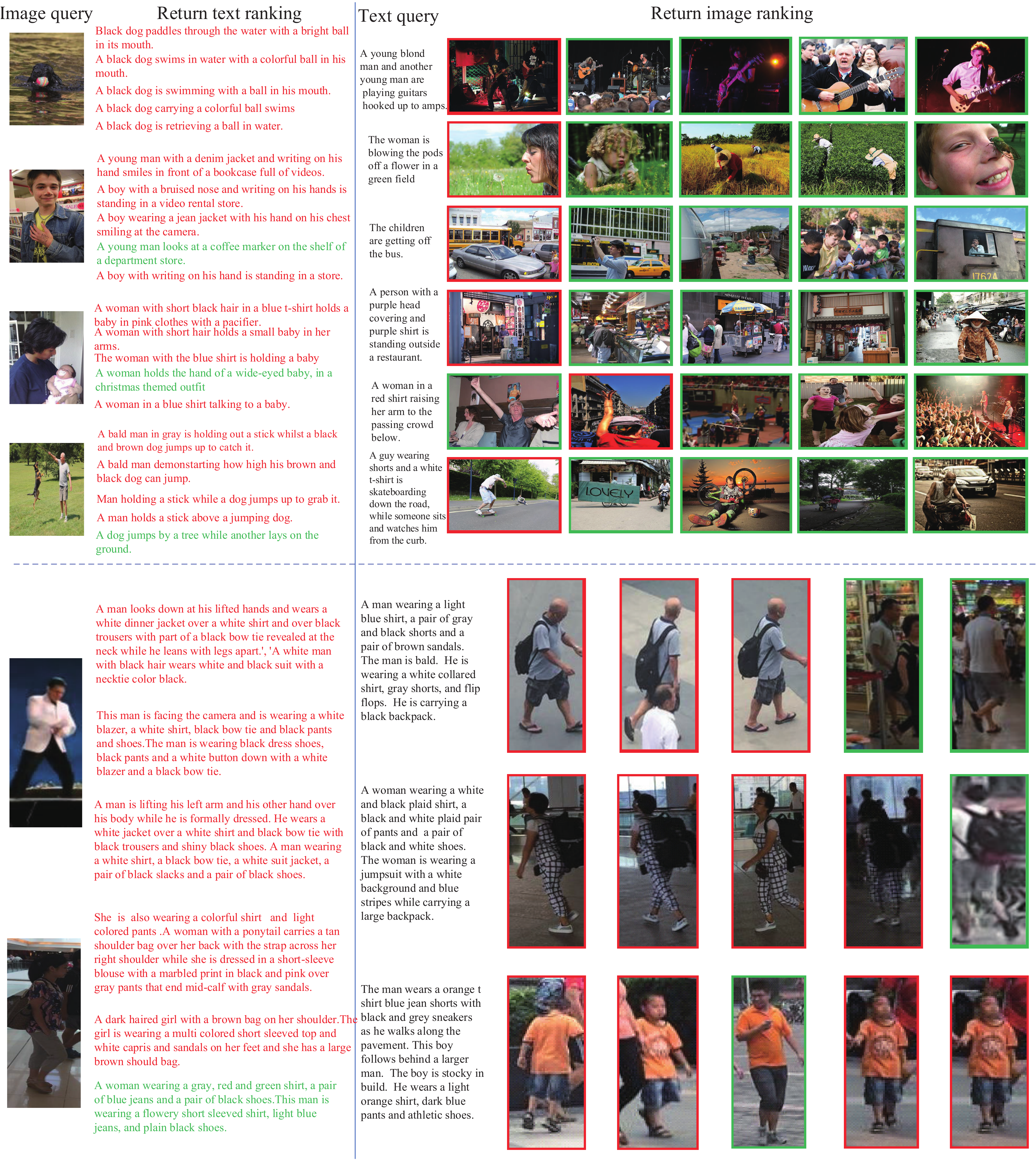} 
\caption{ Qualitative test results on the Flikcr30K and CUHK-PEDES datasets. We report Recall@5 of the ``Image-to-Text'' task and the ``Text-to-Image'' task from left to right. The correct retrieval images or text are in red and a red box, while the failure retrieval are in green. For Flickr30K, each image is described by 5 sentences. Hence, each text query also has a correct retrieved image, but other retrieved images have similar content as described by the sentence. For the CUHK-PEDES dataset, each category has more than one image, thus almost all correct images are retrieved according to the text query. The list is best viewed in color. } \label{Result_Retrieval_show}
\end{figure*}

\section{Conclusion}
\label{Conclusion}

In this work, we explored methods to improve the performance of cross-modal retrieval by integrating information theory and adversarial learning by analyzing the relation between information entropy and modality uncertainty. Based on this relation, we explored two different paradigms to combine information entropy maximization and modality classification in an adversarial manner. Training these two components iteratively reduces feature distribution discrepancies and further the heterogeneity gap. This is beneficial for preserving semantic similarity between cross-modal features by using bi-directional triplet loss and cross-entropy loss. In addition, we also considered the issue of data imbalance, which leads to a biased classifier and affects label classification. KL-divergence is used as an additional loss term to regularize the re-scaled probabilities computed from image features and text features. It is also used to constrain the cross-modal feature projections and is helpful for learning modality-invariant features. The efficacy of the proposed method was demonstrated by thorough experimental results on four well-known datasets using four deep models.

Successfully combining information entropy and adversarial learning depends on the competitive goals between the information entropy predictor and the modality classifier, and this leads to challenging directions worth further investigation. For example, we used instance labels as supervisory information in this work. Then the information entropy loss was computed only based on image modality and text modality. However, retrieval performance depends on the matching of each image-text feature pair. For some large-scale datasets, each category may include a large number of image-text pairs. Thus, it is valuable to make the information entropy loss specific for each category so that the discrepancy between two modalities can be reduced more granularly. Moreover, the problem of data imbalance leads to training a biased label classifier, which is an issue that can also be resolved by training strategies like data augmentation or by using other loss functions, \emph{e.g.} knowledge distillation loss. 
 
In terms of future work, the label-free Shannon information entropy can be used in some unsupervised learning scenarios, and has been used in performing tasks such as semantic segmentation  \cite{vu2019advent}. Examining the application of combining Shannon information entropy with adversarial learning for cross-modal retrieval, we find that Shannon information entropy can be used for multimodal feature learning by estimating the modality uncertainty. It will be promising to explore Shannon entropy further when applied to other kinds of cross-modal feature learning similar to image-text retrieval, such as video-text, audio-video, and audio-text matching which aims at learning modality-invariant representations.

\noindent{
\textbf{Acknowledgments} \\
This work is supported by LIACS MediaLab at Leiden University and China Scholarship Council (CSC No.201703170183). We would like to thank NVIDIA for the donation of GPU cards.}
%
%
%

\begin{spacing}{0.1}
\footnotesize
\bibliographystyle{elsarticle-num-names}
\bibliography{myreference}

\begin{thebibliography}{39}
\providecommand{\natexlab}[1]{#1}
\providecommand{\url}[1]{\texttt{#1}}
\providecommand{\urlprefix}{URL }
\expandafter\ifx\csname urlstyle\endcsname\relax
  \providecommand{\doi}[1]{doi:\discretionary{}{}{}#1}\else
  \providecommand{\doi}[1]{doi:\discretionary{}{}{}\begingroup
  \urlstyle{rm}\url{#1}\endgroup}\fi
\providecommand{\bibinfo}[2]{#2}

\bibitem[{Wang et~al.(2016{\natexlab{a}})Wang, Yin, Wang, Wu, and
  Wang}]{wang2016comprehensive}
\bibinfo{author}{K.~Wang}, \bibinfo{author}{Q.~Yin}, \bibinfo{author}{W.~Wang},
  \bibinfo{author}{S.~Wu}, \bibinfo{author}{L.~Wang}, \bibinfo{title}{A
  comprehensive survey on cross-modal retrieval}, \bibinfo{journal}{arXiv
  preprint arXiv:1607.06215} .

\bibitem[{Li et~al.(2016)Li, Uricchio, Ballan, Bertini, Snoek, and
  Bimbo}]{li2016socializing}
\bibinfo{author}{X.~Li}, \bibinfo{author}{T.~Uricchio},
  \bibinfo{author}{L.~Ballan}, \bibinfo{author}{M.~Bertini},
  \bibinfo{author}{C.~G. Snoek}, \bibinfo{author}{A.~D. Bimbo},
  \bibinfo{title}{Socializing the semantic gap: A comparative survey on image
  tag assignment, refinement, and retrieval}, \bibinfo{journal}{ACM Comput.
  Surv.} \bibinfo{volume}{49}~(\bibinfo{number}{1}) (\bibinfo{year}{2016})
  \bibinfo{pages}{1--39}.

\bibitem[{Angelou et~al.(2019)Angelou, Solachidis, Vretos, and
  Daras}]{angelou2019graph}
\bibinfo{author}{M.~Angelou}, \bibinfo{author}{V.~Solachidis},
  \bibinfo{author}{N.~Vretos}, \bibinfo{author}{P.~Daras},
  \bibinfo{title}{Graph-based multimodal fusion with metric learning for
  multimodal classification}, \bibinfo{journal}{Pattern Recognit.}
  \bibinfo{volume}{95} (\bibinfo{year}{2019}) \bibinfo{pages}{296--307}.

\bibitem[{Baltru{\v{s}}aitis et~al.(2018)Baltru{\v{s}}aitis, Ahuja, and
  Morency}]{baltruvsaitis2018multimodal}
\bibinfo{author}{T.~Baltru{\v{s}}aitis}, \bibinfo{author}{C.~Ahuja},
  \bibinfo{author}{L.-P. Morency}, \bibinfo{title}{Multimodal machine learning:
  A survey and taxonomy}, \bibinfo{journal}{IEEE Trans. Pattern Anal. Mach.
  Intell.} \bibinfo{volume}{41}~(\bibinfo{number}{2}) (\bibinfo{year}{2018})
  \bibinfo{pages}{423--443}.

\bibitem[{Faghri et~al.(2018)Faghri, Fleet, Kiros, and Fidler}]{faghri2017vse}
\bibinfo{author}{F.~Faghri}, \bibinfo{author}{D.~J. Fleet},
  \bibinfo{author}{J.~R. Kiros}, \bibinfo{author}{S.~Fidler},
  \bibinfo{title}{Vse++: Improving visual-semantic embeddings with hard
  negatives}, in: \bibinfo{booktitle}{Proc. BMVC}, \bibinfo{pages}{1--10},
  \bibinfo{year}{2018}.

\bibitem[{Wang et~al.(2017)Wang, Yang, Xu, Hanjalic, and
  Shen}]{wang2017adversarial}
\bibinfo{author}{B.~Wang}, \bibinfo{author}{Y.~Yang}, \bibinfo{author}{X.~Xu},
  \bibinfo{author}{A.~Hanjalic}, \bibinfo{author}{H.~T. Shen},
  \bibinfo{title}{Adversarial cross-modal retrieval}, in:
  \bibinfo{booktitle}{Proc. ACM MM}, \bibinfo{pages}{154--162},
  \bibinfo{year}{2017}.

\bibitem[{Liu et~al.(2019)Liu, Guo, Liu, Bakker, and Lew}]{liu2019cyclematch}
\bibinfo{author}{Y.~Liu}, \bibinfo{author}{Y.~Guo}, \bibinfo{author}{L.~Liu},
  \bibinfo{author}{E.~M. Bakker}, \bibinfo{author}{M.~S. Lew},
  \bibinfo{title}{CycleMatch: A cycle-consistent embedding network for
  image-text matching}, \bibinfo{journal}{Pattern Recognit.}
  \bibinfo{volume}{93} (\bibinfo{year}{2019}) \bibinfo{pages}{365--379}.

\bibitem[{Shannon(1948)}]{shannon1948mathematical}
\bibinfo{author}{C.~E. Shannon}, \bibinfo{title}{A mathematical theory of
  communication}, \bibinfo{journal}{Bell system technical journal}
  \bibinfo{volume}{27}~(\bibinfo{number}{3}) (\bibinfo{year}{1948})
  \bibinfo{pages}{379--423}.

\bibitem[{Mhiri et~al.(2019)Mhiri, Desrosiers, and Cheriet}]{mhiri2019word}
\bibinfo{author}{M.~Mhiri}, \bibinfo{author}{C.~Desrosiers},
  \bibinfo{author}{M.~Cheriet}, \bibinfo{title}{Word spotting and recognition
  via a joint deep embedding of image and text}, \bibinfo{journal}{Pattern
  Recognit.} \bibinfo{volume}{88} (\bibinfo{year}{2019})
  \bibinfo{pages}{312--320}.

\bibitem[{Wang et~al.(2020)Wang, Wang, He, Gao, and Tian}]{wang2020joint}
\bibinfo{author}{D.~Wang}, \bibinfo{author}{Q.~Wang}, \bibinfo{author}{L.~He},
  \bibinfo{author}{X.~Gao}, \bibinfo{author}{Y.~Tian}, \bibinfo{title}{Joint
  and Individual Matrix Factorization Hashing for Large-scale Cross-modal
  Retrieval}, \bibinfo{journal}{Pattern Recognit.}  (\bibinfo{year}{2020})
  \bibinfo{pages}{107479}.

\bibitem[{Wu et~al.(2020)Wu, Jing, Wu, Ji, Dong, Luo, Huang, and
  Wang}]{wu2020modality}
\bibinfo{author}{F.~Wu}, \bibinfo{author}{X.-Y. Jing}, \bibinfo{author}{Z.~Wu},
  \bibinfo{author}{Y.~Ji}, \bibinfo{author}{X.~Dong}, \bibinfo{author}{X.~Luo},
  \bibinfo{author}{Q.~Huang}, \bibinfo{author}{R.~Wang},
  \bibinfo{title}{Modality-specific and shared generative adversarial network
  for cross-modal retrieval}, \bibinfo{journal}{Pattern Recognit.}
  (\bibinfo{year}{2020}) \bibinfo{pages}{107335}.

\bibitem[{Zhang and Lu(2018)}]{zhang2018deep}
\bibinfo{author}{Y.~Zhang}, \bibinfo{author}{H.~Lu}, \bibinfo{title}{Deep
  cross-modal projection learning for image-text matching}, in:
  \bibinfo{booktitle}{Proc. ECCV}, \bibinfo{pages}{686--701},
  \bibinfo{year}{2018}.

\bibitem[{Vu et~al.(2019)Vu, Jain, Bucher, Cord, and P{\'e}rez}]{vu2019advent}
\bibinfo{author}{T.-H. Vu}, \bibinfo{author}{H.~Jain},
  \bibinfo{author}{M.~Bucher}, \bibinfo{author}{M.~Cord},
  \bibinfo{author}{P.~P{\'e}rez}, \bibinfo{title}{Advent: Adversarial entropy
  minimization for domain adaptation in semantic segmentation}, in:
  \bibinfo{booktitle}{Proc. IEEE CVPR}, \bibinfo{pages}{2517--2526},
  \bibinfo{year}{2019}.

\bibitem[{Chen et~al.(2019)Chen, Pu, Liu, Bakker, and Lew}]{chen2019domain}
\bibinfo{author}{W.~Chen}, \bibinfo{author}{N.~Pu}, \bibinfo{author}{Y.~Liu},
  \bibinfo{author}{E.~Bakker}, \bibinfo{author}{M.~Lew}, \bibinfo{title}{Domain
  Uncertainty Based On Information Theory for Cross-Modal Hash Retrieval}, in:
  \bibinfo{booktitle}{Proc. IEEE ICME}, \bibinfo{pages}{43--48},
  \bibinfo{year}{2019}.

\bibitem[{He et~al.(2016)He, Zhang, Ren, and Sun}]{he2016deep}
\bibinfo{author}{K.~He}, \bibinfo{author}{X.~Zhang}, \bibinfo{author}{S.~Ren},
  \bibinfo{author}{J.~Sun}, \bibinfo{title}{Deep residual learning for image
  recognition}, in: \bibinfo{booktitle}{Proc. IEEE CVPR},
  \bibinfo{pages}{770--778}, \bibinfo{year}{2016}.

\bibitem[{Howard et~al.(2017)Howard, Zhu, Chen, Kalenichenko, Wang, Weyand,
  Andreetto, and Adam}]{howard2017mobilenets}
\bibinfo{author}{A.~G. Howard}, \bibinfo{author}{M.~Zhu},
  \bibinfo{author}{B.~Chen}, \bibinfo{author}{D.~Kalenichenko},
  \bibinfo{author}{W.~Wang}, \bibinfo{author}{T.~Weyand},
  \bibinfo{author}{M.~Andreetto}, \bibinfo{author}{H.~Adam},
  \bibinfo{title}{Mobilenets: Efficient convolutional neural networks for
  mobile vision applications}, \bibinfo{journal}{arXiv preprint
  arXiv:1704.04861} .

\bibitem[{Graves et~al.(2005)Graves, Fern{\'a}ndez, and
  Schmidhuber}]{graves2005bidirectional}
\bibinfo{author}{A.~Graves}, \bibinfo{author}{S.~Fern{\'a}ndez},
  \bibinfo{author}{J.~Schmidhuber}, \bibinfo{title}{Bidirectional LSTM networks
  for improved phoneme classification and recognition}, in:
  \bibinfo{booktitle}{Proc. ICANN}, \bibinfo{organization}{Springer},
  \bibinfo{pages}{799--804}, \bibinfo{year}{2005}.

\bibitem[{Zhong et~al.(2020)Zhong, Chen, Min, and Xia}]{zhong2020novel}
\bibinfo{author}{F.~Zhong}, \bibinfo{author}{Z.~Chen},
  \bibinfo{author}{G.~Min}, \bibinfo{author}{F.~Xia}, \bibinfo{title}{A novel
  strategy to balance the results of cross-modal hashing},
  \bibinfo{journal}{Pattern Recognit.} \bibinfo{volume}{107}
  (\bibinfo{year}{2020}) \bibinfo{pages}{107523}.

\bibitem[{Guo et~al.(2017)Guo, Pleiss, Sun, and
  Weinberger}]{guo2017calibration}
\bibinfo{author}{C.~Guo}, \bibinfo{author}{G.~Pleiss},
  \bibinfo{author}{Y.~Sun}, \bibinfo{author}{K.~Q. Weinberger},
  \bibinfo{title}{On Calibration of Modern Neural Networks}, in:
  \bibinfo{booktitle}{Proc. ICML}, \bibinfo{pages}{1321--1330},
  \bibinfo{year}{2017}.

\bibitem[{Hodosh et~al.(2013)Hodosh, Young, and
  Hockenmaier}]{hodosh2013framing}
\bibinfo{author}{M.~Hodosh}, \bibinfo{author}{P.~Young},
  \bibinfo{author}{J.~Hockenmaier}, \bibinfo{title}{Framing image description
  as a ranking task: Data, models and evaluation metrics},
  \bibinfo{journal}{Journal of Artificial Intelligence Research}
  \bibinfo{volume}{47} (\bibinfo{year}{2013}) \bibinfo{pages}{853--899}.

\bibitem[{Young et~al.(2014)Young, Lai, Hodosh, and
  Hockenmaier}]{young2014image}
\bibinfo{author}{P.~Young}, \bibinfo{author}{A.~Lai},
  \bibinfo{author}{M.~Hodosh}, \bibinfo{author}{J.~Hockenmaier},
  \bibinfo{title}{From image descriptions to visual denotations: New similarity
  metrics for semantic inference over event descriptions},
  \bibinfo{journal}{Trans. Association for Computational Linguistics}
  \bibinfo{volume}{2} (\bibinfo{year}{2014}) \bibinfo{pages}{67--78}.

\bibitem[{Lin et~al.(2014)Lin, Maire, Belongie, Hays, Perona, Ramanan,
  Doll{\'a}r, and Zitnick}]{lin2014microsoft}
\bibinfo{author}{T.-Y. Lin}, \bibinfo{author}{M.~Maire},
  \bibinfo{author}{S.~Belongie}, \bibinfo{author}{J.~Hays},
  \bibinfo{author}{P.~Perona}, \bibinfo{author}{D.~Ramanan},
  \bibinfo{author}{P.~Doll{\'a}r}, \bibinfo{author}{C.~L. Zitnick},
  \bibinfo{title}{Microsoft coco: Common objects in context}, in:
  \bibinfo{booktitle}{Proc. ECCV}, \bibinfo{pages}{740--755},
  \bibinfo{year}{2014}.

\bibitem[{Li et~al.(2017{\natexlab{a}})Li, Xiao, Li, Zhou, Yue, and
  Wang}]{li2017person}
\bibinfo{author}{S.~Li}, \bibinfo{author}{T.~Xiao}, \bibinfo{author}{H.~Li},
  \bibinfo{author}{B.~Zhou}, \bibinfo{author}{D.~Yue},
  \bibinfo{author}{X.~Wang}, \bibinfo{title}{Person search with natural
  language description}, in: \bibinfo{booktitle}{Proc. IEEE CVPR},
  \bibinfo{pages}{1970--1979}, \bibinfo{year}{2017}{\natexlab{a}}.

\bibitem[{Simonyan and Zisserman(2014)}]{simonyan2014very}
\bibinfo{author}{K.~Simonyan}, \bibinfo{author}{A.~Zisserman},
  \bibinfo{title}{Very deep convolutional networks for large-scale image
  recognition}, \bibinfo{journal}{arXiv preprint arXiv:1409.1556} .

\bibitem[{Wang et~al.(2016{\natexlab{b}})Wang, Li, and
  Lazebnik}]{wang2016learning}
\bibinfo{author}{L.~Wang}, \bibinfo{author}{Y.~Li},
  \bibinfo{author}{S.~Lazebnik}, \bibinfo{title}{Learning deep
  structure-preserving image-text embeddings}, in: \bibinfo{booktitle}{Proc.
  IEEE CVPR}, \bibinfo{pages}{5005--5013}, \bibinfo{year}{2016}{\natexlab{b}}.

\bibitem[{Mao et~al.(2015)Mao, Xu, Yang, Wang, Huang, and Yuille}]{mao2014deep}
\bibinfo{author}{J.~Mao}, \bibinfo{author}{W.~Xu}, \bibinfo{author}{Y.~Yang},
  \bibinfo{author}{J.~Wang}, \bibinfo{author}{Z.~Huang},
  \bibinfo{author}{A.~Yuille}, \bibinfo{title}{Deep Captioning with Multimodal
  Recurrent Neural Networks (m-RNN)}, in: \bibinfo{booktitle}{Proc. ICLR},
  \bibinfo{year}{2015}.

\bibitem[{Lev et~al.(2016)Lev, Sadeh, Klein, and Wolf}]{lev2016rnn}
\bibinfo{author}{G.~Lev}, \bibinfo{author}{G.~Sadeh},
  \bibinfo{author}{B.~Klein}, \bibinfo{author}{L.~Wolf}, \bibinfo{title}{Rnn
  fisher vectors for action recognition and image annotation}, in:
  \bibinfo{booktitle}{Proc. ECCV}, \bibinfo{pages}{833--850},
  \bibinfo{year}{2016}.

\bibitem[{Dong et~al.(2018)Dong, Li, and Snoek}]{dong2018predicting}
\bibinfo{author}{J.~Dong}, \bibinfo{author}{X.~Li}, \bibinfo{author}{C.~G.
  Snoek}, \bibinfo{title}{Predicting visual features from text for image and
  video caption retrieval}, \bibinfo{journal}{IEEE Trans. Multimedia}
  \bibinfo{volume}{20}~(\bibinfo{number}{12}) (\bibinfo{year}{2018})
  \bibinfo{pages}{3377--3388}.

\bibitem[{Huang et~al.(2017)Huang, Wang, and Wang}]{huang2017instance}
\bibinfo{author}{Y.~Huang}, \bibinfo{author}{W.~Wang},
  \bibinfo{author}{L.~Wang}, \bibinfo{title}{Instance-aware image and sentence
  matching with selective multimodal lstm}, in: \bibinfo{booktitle}{Proc. IEEE
  CVPR}, \bibinfo{pages}{2310--2318}, \bibinfo{year}{2017}.

\bibitem[{Liu et~al.(2017)Liu, Guo, Bakker, and Lew}]{liu2017learning}
\bibinfo{author}{Y.~Liu}, \bibinfo{author}{Y.~Guo}, \bibinfo{author}{E.~M.
  Bakker}, \bibinfo{author}{M.~S. Lew}, \bibinfo{title}{Learning a recurrent
  residual fusion network for multimodal matching}, in:
  \bibinfo{booktitle}{Proc. IEEE ICCV}, \bibinfo{pages}{4107--4116},
  \bibinfo{year}{2017}.

\bibitem[{Wang et~al.(2019)Wang, Guo, Xu, Zhuo, and Wang}]{wang2019cross}
\bibinfo{author}{S.~Wang}, \bibinfo{author}{D.~Guo}, \bibinfo{author}{X.~Xu},
  \bibinfo{author}{L.~Zhuo}, \bibinfo{author}{M.~Wang},
  \bibinfo{title}{Cross-Modality Retrieval by Joint Correlation Learning},
  \bibinfo{journal}{ACM Trans. Multimedia Comput. Commun. Appl.}
  \bibinfo{volume}{15}~(\bibinfo{number}{2s}) (\bibinfo{year}{2019})
  \bibinfo{pages}{56}.

\bibitem[{Sarafianos et~al.(2019)Sarafianos, Xu, and
  Kakadiaris}]{sarafianos2019adversarial}
\bibinfo{author}{N.~Sarafianos}, \bibinfo{author}{X.~Xu},
  \bibinfo{author}{I.~A. Kakadiaris}, \bibinfo{title}{Adversarial
  representation learning for text-to-image matching}, in:
  \bibinfo{booktitle}{Proc. IEEE ICCV}, \bibinfo{pages}{5814--5824},
  \bibinfo{year}{2019}.

\bibitem[{Nam et~al.(2017)Nam, Ha, and Kim}]{nam2017dual}
\bibinfo{author}{H.~Nam}, \bibinfo{author}{J.-W. Ha}, \bibinfo{author}{J.~Kim},
  \bibinfo{title}{Dual attention networks for multimodal reasoning and
  matching}, in: \bibinfo{booktitle}{Proc. IEEE CVPR},
  \bibinfo{pages}{299--307}, \bibinfo{year}{2017}.

\bibitem[{Zheng et~al.(2020)Zheng, Zheng, Garrett, Yang, Xu, and
  Shen}]{zheng2017dual}
\bibinfo{author}{Z.~Zheng}, \bibinfo{author}{L.~Zheng},
  \bibinfo{author}{M.~Garrett}, \bibinfo{author}{Y.~Yang},
  \bibinfo{author}{M.~Xu}, \bibinfo{author}{Y.-D. Shen},
  \bibinfo{title}{Dual-Path Convolutional Image-Text Embeddings with Instance
  Loss}, \bibinfo{journal}{ACM Trans. Multimedia Comput. Commun. Appl.}
  \bibinfo{volume}{16}~(\bibinfo{number}{2}) (\bibinfo{year}{2020})
  \bibinfo{pages}{1--23}.

\bibitem[{Li et~al.(2017{\natexlab{b}})Li, Xiao, Li, Yang, and
  Wang}]{li2017identity}
\bibinfo{author}{S.~Li}, \bibinfo{author}{T.~Xiao}, \bibinfo{author}{H.~Li},
  \bibinfo{author}{W.~Yang}, \bibinfo{author}{X.~Wang},
  \bibinfo{title}{Identity-aware textual-visual matching with latent
  co-attention}, in: \bibinfo{booktitle}{Proc. IEEE ICCV},
  \bibinfo{pages}{1890--1899}, \bibinfo{year}{2017}{\natexlab{b}}.

\bibitem[{Chen et~al.(2018)Chen, Li, Liu, Shen, Shao, Yuan, and
  Wang}]{chen2018improving}
\bibinfo{author}{D.~Chen}, \bibinfo{author}{H.~Li}, \bibinfo{author}{X.~Liu},
  \bibinfo{author}{Y.~Shen}, \bibinfo{author}{J.~Shao},
  \bibinfo{author}{Z.~Yuan}, \bibinfo{author}{X.~Wang},
  \bibinfo{title}{Improving deep visual representation for person
  re-identification by global and local image-language association}, in:
  \bibinfo{booktitle}{Proc. ECCV}, \bibinfo{pages}{54--70},
  \bibinfo{year}{2018}.

\bibitem[{Niu et~al.(2020)Niu, Huang, Ouyang, and Wang}]{niu2019improving}
\bibinfo{author}{K.~Niu}, \bibinfo{author}{Y.~Huang},
  \bibinfo{author}{W.~Ouyang}, \bibinfo{author}{L.~Wang},
  \bibinfo{title}{Improving description-based person re-identification by
  multi-granularity image-text alignments}, \bibinfo{journal}{IEEE Trans. Image
  Process.} \bibinfo{volume}{29} (\bibinfo{year}{2020})
  \bibinfo{pages}{5542--5556}.

\bibitem[{Klein et~al.(2015)Klein, Lev, Sadeh, and Wolf}]{klein2015associating}
\bibinfo{author}{B.~Klein}, \bibinfo{author}{G.~Lev},
  \bibinfo{author}{G.~Sadeh}, \bibinfo{author}{L.~Wolf},
  \bibinfo{title}{Associating neural word embeddings with deep image
  representations using fisher vectors}, in: \bibinfo{booktitle}{Proc. IEEE
  CVPR}, \bibinfo{pages}{4437--4446}, \bibinfo{year}{2015}.

\bibitem[{Ganin and Lempitsky(2015)}]{ganin2015unsupervised}
\bibinfo{author}{Y.~Ganin}, \bibinfo{author}{V.~Lempitsky},
  \bibinfo{title}{Unsupervised domain adaptation by backpropagation}, in:
  \bibinfo{booktitle}{Proc. ICML}, \bibinfo{pages}{1180--1189},
  \bibinfo{year}{2015}.

\end{thebibliography}
\end{spacing}
\end{document}